%% file: main.tex
\documentclass[final]{dynafront2026} 

\title[Functional Reachability at Extreme Sparsity]{When Sparsity Breaks Connectivity: Functional Reachability and Guided Topology Dynamics in Extreme Sparse Training}

\optauthor{%
\Name{Mohammad Pasande}\thanks{Equal contribution.}$^{\rm 1}$ \Email{m.pasande@utwente.nl}\\
\Name{Mateusz Jaworski}\footnotemark[\value{footnote}]$^{\rm 1}$ \Email{m.z.jaworski@alumnus.utwente.nl}\\
\Name{Jeroen Rook}$^{\rm 2}$ \Email{jeroen.rook@uni-paderborn.de}\\
\Name{Elena Mocanu}$^{\rm 1}$ \Email{e.mocanu@utwente.nl}\\
\addr $^{\rm 1}$University of Twente, Enschede, Netherlands\\
$^{\rm 2}$Paderborn University, Paderborn, Germany
}

\usepackage{booktabs}
\usepackage{multirow}
\usepackage{xcolor}
\usepackage{microtype}
\usepackage{enumitem}

\include{commenting_macros}

\begin{document}

\maketitle

\begin{abstract}%
Dynamic Sparse Training (DST) periodically prunes and regrows connections, inducing a discrete dynamical mechanism over network topologies.
At extreme sparsity ($\ge\!99\%$), this dynamic tends to destroy input--output information pathways, and performance with standard magnitude/random updates degrades sharply.
We formalize \emph{forward}, \emph{backward}, and \emph{functional} reachability on layered sparse graphs, derive closed-form expectations for reachable neurons and edges, and show that Monte Carlo realizations match the theory.
Guided by this reachability-based structure, we bias pruning toward non-functional edges and regrowth toward functionally consistent candidates.
We introduce this reachability-driven modification of topology evolution as \emph{Reachability-guided} in DST, referred to as \emph{guided DST}.
On CIFAR-10/100 and Tiny ImageNet, guided DST recovers large accuracy gaps versus unguided DST (up to $+14.4$ points at $99.9\%$ sparsity on CIFAR-100), and on MLP-Mixer guided DST remains trainable at $99.99\%$ sparsity while unguided random regrowth degrades sharply.
These results cast extreme-sparse training as a connectivity-preserving dynamical process. Our implementation is publicly available at \href{https://github.com/mzjaworski/ea_dst}{GitHub}.
\end{abstract}

\section{Introduction}
\label{sec:intro}

Sparse neural networks offer the potential to reduce memory and computational requirements, but training under a fixed sparsity constraint requires optimizing both the network parameters and its connectivity structure~\citep{mocanu2018scalable, hoefler2021sparsity}.
Dynamic Sparse Training (DST) methods such as SET~\citep{mocanu2018scalable} and RigL~\citep{evci2020rigging} address this by iterative updates with discrete topology mutations (remove then regrow), yielding a hybrid continuous--discrete dynamical system on the joint space of weights and binary connectivity masks~\citep{ji2024advancing}.

Empirically, DST can retain strong performance across a broad range of sparsity levels, but its reliability becomes increasingly sensitive to the allocation and evolution of the remaining connections as density decreases~\citep{evci2020rigging,liu2021itop}. Pushing sparsity to more extreme levels is of particular interest because it probes how far neural networks can reduce their connection allocation while remaining trainable. However, beyond a certain sparsity level, performance can deteriorate abruptly rather than gradually. With very few active connections, individual layers may become severely under-connected or effectively collapse, disrupting information and gradient propagation and limiting the expressive capacity of the resulting sparse network~\citep{tanaka2020pruning,li2025pushing}. We refer to this highly sparse regime, where maintaining viable connectivity becomes a central challenge, as the \emph{extreme-sparsity regime}.

A complementary line of work studies sparse neural networks through their graph-theoretic properties and the organization and evolution of their active connections~\citep{liu2020topological}. Connectivity-aware regularization, such as CoNNect, further promotes input--output connectivity and avoids structural failures such as layer collapse during pruning~\citep{franssen2025connect}. However, these approaches do not classify individual active edges according to whether they lie on one complete input--output path, nor do they perform DST. Our work formalizes this distinction through forward, backward, and functional reachability and uses it to guide both pruning and regrowth. 
We make three main contributions:

\begin{enumerate}[itemsep=0pt, parsep=0pt] 
\item A layered-graph formulation of forward, backward, and functional
reachability, with analytical approximations for reachable vertices and edges.
\item Empirical evidence that functional-reachability collapse aligns with the sparsity regime where unguided DST performance deteriorates.
\item A reachability-guided DST method that preserves functional connectivity,
improving performance at sparsities $\geq 99.75\%$ and transferring to
MLP-Mixer up to $99.99\%$ sparsity.
\end{enumerate}
Further details on the formalism, algorithms, and experiments are presented in~\citet{Jaworski2026}.

\section{Reachability Formalism}
\label{sec:theory}

Classical Erd\H{o}s--R\'enyi (ER) random graphs exhibit connectivity thresholds as edge density decreases~\citep{erdos1960evolution}. In sparse neural networks, connectivity alone is not enough: learning needs bidirectional pathways from inputs to outputs~\citep{franssen2025connect,liu2020topological}.
We model a feed-forward neural network as a layered graph $G=(V,E)$ with
$V=\bigcup_{l=0}^{L}V_l$, candidate edges
$\mathcal{U}^{e}=\bigcup_{l=1}^{L}(V_{l-1}\times V_l)$, and $E\subseteq\mathcal{U}^{e}$.
Edges between $V_{l-1}$ and $V_l$ appear independently with probability $p_l\in[0,1]$.

\begin{figure}[ht]
\centering
\setlength{\tabcolsep}{4pt}%
\begin{tabular}{@{}cc@{}}
\includegraphics[width=0.38\linewidth]{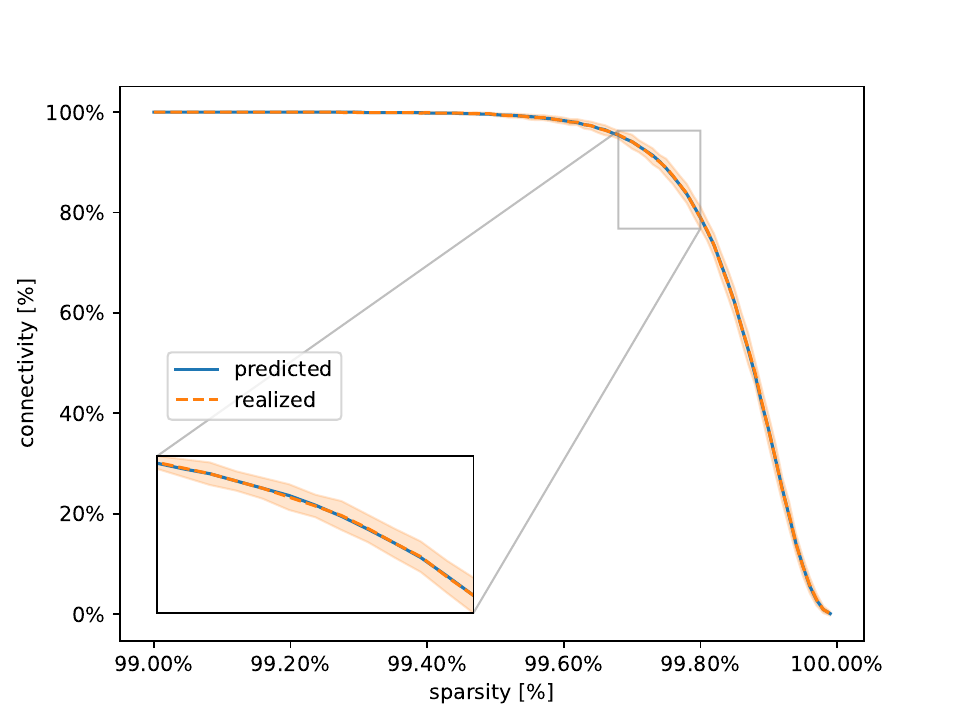} &
\includegraphics[width=0.38\linewidth]{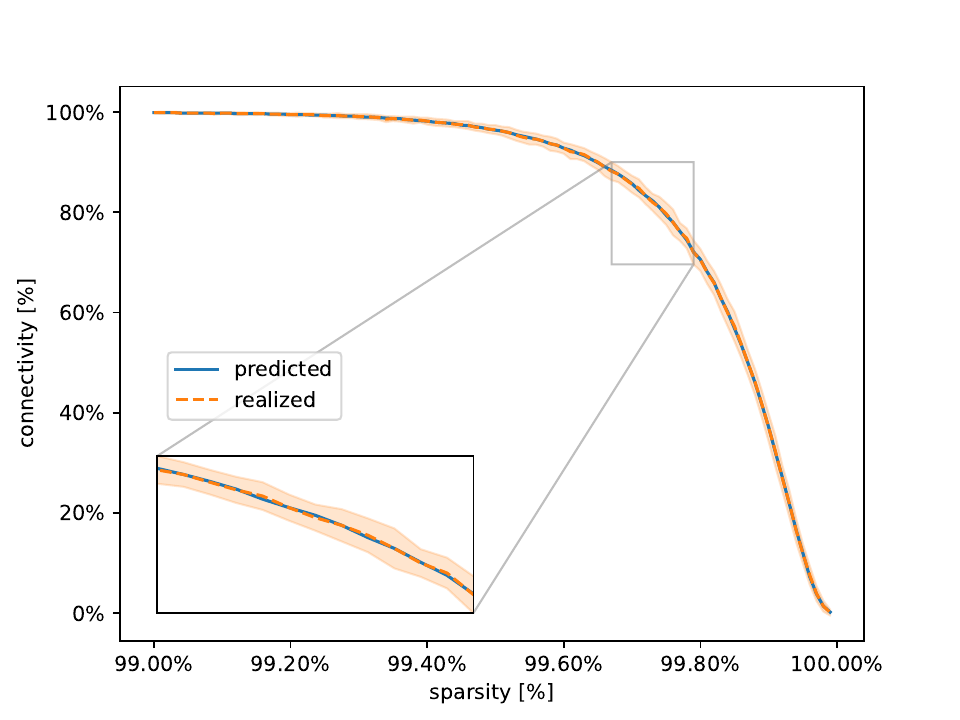} \\[-0.1em]
\footnotesize (a) Layer~1 (functional) &
\footnotesize (b) Layer~2 (functional)
\end{tabular}
\caption{Functional edge reachability under an ER initialization for an extreme sparse MLP. The predicted (solid) vs. realizated (dashed) are near-perfect until $\sim\!99.6$--$99.7\%$ sparsity. The full grids can be seen in Appendix~\ref{app:theory_mc}.}
\label{fig:reachability}
\end{figure}

\subsection{Reachability sets}\label{sec:theory-reach-set}
Forward- and backward-reachable vertices are defined recursively from the input and output layers:
\begin{align}
R^{\mathrm{fwd},v}_{0}&=V_0,&
R^{\mathrm{fwd},v}_{l}
&=\bigl\{v\in V_l \mid \exists\,u\in R^{\mathrm{fwd},v}_{l-1}:\,(u,v)\in E\bigr\},\\
R^{\mathrm{bwd},v}_{L}&=V_L,&
R^{\mathrm{bwd},v}_{l}
&=\bigl\{u\in V_l \mid \exists\,v\in R^{\mathrm{bwd},v}_{l+1}:\,(u,v)\in E\bigr\}.
\end{align}
Functional vertices are forward- and backward-reachable intersection, $R^{\mathrm{fcnl},v}_{l}=R^{\mathrm{fwd},v}_{l}\cap R^{\mathrm{bwd},v}_{l}$,
i.e., neurons on complete input--output paths that can carry both activations and
gradient updates.
The active \emph{functional} edges used in Section~\ref{sec:method} are
\begin{equation}
R^{\mathrm{fcnl},e}_{l}
=\bigl\{(u,v)\in E_l \mid u\in R^{\mathrm{fwd},v}_{l-1},\;
v\in R^{\mathrm{bwd},v}_{l}\bigr\}.
\end{equation}
Full endpoint-consistent edge sets and the remaining derivations are in
Appendix~\ref{app:theory_formalism}.
Under independent Bernoulli edges, mean-field expectations satisfy, in
particular:
\begin{align}
\mathbb{E}\!\bigl[|R^{\mathrm{fcnl},e}_{l}|\bigr]
&\approx
p_l\,\mathbb{E}\!\bigl[|R^{\mathrm{fwd},v}_{l-1}|\bigr]\,
\mathbb{E}\!\bigl[|R^{\mathrm{bwd},v}_{l}|\bigr],
\label{eq:E-fcnl-e}\\
\mathbb{E}\!\bigl[|R^{\mathrm{fcnl},v}_{l}|\bigr]
&\approx
\frac{\mathbb{E}[|R^{\mathrm{fwd},v}_{l}|]\,\mathbb{E}[|R^{\mathrm{bwd},v}_{l}|]}{|V_l|}.
\label{eq:E-fcnl-v}
\end{align}
Functional connectivity is limited by the overlap of the two directions: once
either forward or backward reachability thins, the functional fraction falls
multiplicatively rather than additively, predicting sharp connectivity cliffs
at extreme sparsity.

\subsection{Validation}
Evaluating~\eqref{eq:E-fcnl-e}--\eqref{eq:E-fcnl-v} and the forward/backward
recursions in Appendix~\ref{app:theory_formalism} on our MLP widths and
comparing to Monte Carlo realizations (Figure~\ref{fig:reachability}) shows close
agreement.
Under ER initialization, functional connectivity remains near-perfect until a
sharp cliff around $\sim\!99.6$--$99.7\%$ sparsity; under uniform
initialization, it decays earlier and more gradually with deeper-layer
bottlenecks.
Thus the mean-field analysis correctly identifies the extreme-sparsity regime
in which input--output pathways are expected to break for fixed random masks.

\section{Guided Topology Dynamics}
\label{sec:method}

DST has three main steps:
\textbf{(i)~initialization} of a sparse mask at a target sparsity,
\textbf{(ii)~removal} of $k_l$ active edges per layer, and
\textbf{(iii)~regrowth} of $k_l$ previously inactive edges.
Continuous weight updates are left unchanged; we modify only the removal and
regrowth operators using the reachability sets of Section~\ref{sec:theory}.

\paragraph{Guided removal.}
For each active edge $(u,v)\in E_l$, define the removal score
\begin{equation}
\pi^{(l)}_{uv}
=
\left(
\mathbf{1}\!\left[(u,v)\in R^{\mathrm{fcnl},e}_l\right],
\,
\left|w^{(l)}_{uv}\right|
\right).
\label{eq:guided-removal}
\end{equation}
Lexicographic order removes non-functional edges before functional ones, and
within each group prefers smaller magnitudes.
If fewer than $k_l$ non-functional edges exist, the remainder are taken from
$R^{\mathrm{fcnl},e}_l$.
Reachability is computed once per topology update and reused across layers
(two sparse structural passes; small overhead vs.\ a training step).

\paragraph{Guided regrowth.}
After removal, inactive candidates are
$\mathcal{I}_l=(V_{l-1}\times V_l)\setminus E_l^{-}$.
Functionally consistent regrowth targets
\begin{equation}
R^{\mathrm{guide},e}_l
=
\left\{
(u,v)\in\mathcal{I}_l
\;\middle|\;
u\in R^{\mathrm{fwd},v}_{l-1}(E^{-}),\;
v\in R^{\mathrm{bwd},v}_{l}(E^{-})
\right\},
\label{eq:guide-set}
\end{equation}
i.e., edges that close an input--output path.
We sample $k_l$ replacements with strength $\beta{=}0.99$ on this set
(fallback when the set is empty, that is described Appendix~\ref{app:guided_regrowth}),
steering the topology toward a functionally active subgraph.

\paragraph{Guided DST.}
Combining the standard DST loop with the removal and regrowth rules above
yields \emph{guided DST}: after mask initialization, each topology update
removes and regrows edges under reachability bias.
Algorithm~\ref{alg:guided-dst} summarizes one such update.
\vspace{-0.1cm}
\begin{algorithm2e}[ht]
\caption{Reachability-guided DST topology update}
\label{alg:guided-dst}
\SetAlgoLined
\KwIn{weights $\{W_l\}$, active edges $\{E_l\}$, prune counts $\{k_l\}$, $\beta$}
Compute $R^{\mathrm{fwd},v}$, $R^{\mathrm{bwd},v}$, $R^{\mathrm{fcnl},e}$ once\;
\For{each masked layer $l$}{
  \textbf{Remove:} drop the $k_l$ edges with smallest $\pi^{(l)}_{uv}$
  (non-functional first, then by $|w_{uv}|$); obtain $E_l^{-}$\;
  \textbf{Regrow:} sample $k_l$ edges from
  $\mathcal{I}_l=(V_{l-1}\times V_l)\setminus E_l^{-}$ with mass
  $\beta$ on $R^{\mathrm{guide},e}_l$ (or its fallback) and $1{-}\beta$
  elsewhere\;
}
\end{algorithm2e}

\begin{figure}[h]
\centering
\setlength{\tabcolsep}{4pt}%
\begin{tabular}{@{}cc@{}}
\includegraphics[width=0.48\linewidth]{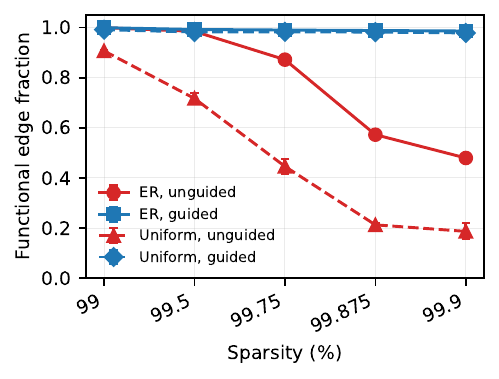} &
\includegraphics[width=0.48\linewidth]{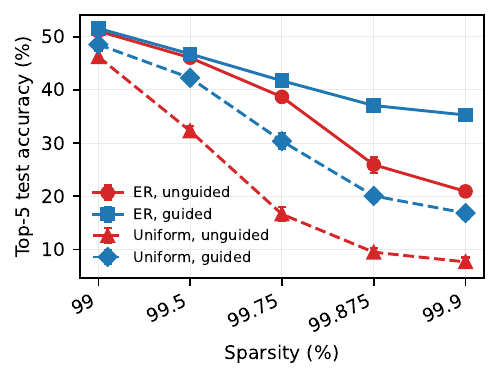} \\[-0.1em]
(a) Functional edge fraction &
(b) Top-5 test accuracy
\end{tabular}
\caption{Empirical reachability and accuracy after training MLPs with DST on
CIFAR-100 (final epoch, five seeds).}
\label{fig:reachability-dst}
\end{figure}

\section{Experiments}
\label{sec:expts}
We evaluate reachability-guided DST against the unguided Magnitude/Random baseline on $4$-layer MLPs trained on CIFAR-10, CIFAR-100~\citep{krizhevsky2009learning},
and Tiny ImageNet~\citep{le2015tiny}, using both ER and uniform mask initialization at sparsities
ranging from $99\%$ to $99.9\%$. We further evaluate transfer to an MLP-Mixer~\citep{tolstikhin2021mlp} on CIFAR-100 under ER initialization, up to $99.99\%$ sparsity. Protocols and architectures are in Appendix~\ref{app:setup}.

Section~\ref{sec:theory} analyzes a fixed sparse topology; DST, however,
{evolves} the topology during training.
We therefore log, at every epoch, the fraction of active edges that are
functionally reachable, averaged over masked layers.
Figure~\ref{fig:reachability-dst} compares unguided DST (magnitude pruning,
random regrowth) with guided DST on CIFAR-100 MLPs.
Under ER initialization, both protocols maintain near-perfect functional
fractions at $99\%$--$99.5\%$ sparsity; from $99.75\%$ onward, unguided DST
progressively loses functional connectivity, reaching $48\%$ functional
edges at $99.9\%$, whereas guided DST remains above $ 98\%$.
A similar separation occurs under uniform initialization, where the functional
fraction of unguided DST falls to $19\%$ at $99.9\%$.
Extended training curves are in
Appendix~\ref{app:reachability}.

\begin{table}[ht]
\centering
\scriptsize
\setlength{\tabcolsep}{3pt}
\renewcommand{\arraystretch}{0.95}
\caption{Top-5 test accuracy ($\%$, mean$\pm$std over five seeds) of
$4$-layer MLPs trained at $99.9\%$ sparsity with unguided Magnitude/Random
versus reachability-guided.}
\label{tab:mlp}
\begin{tabular}{llccc}
\toprule
Initialization & Dataset & Unguided Mag./Rand. & Guided Mag./Rand. & $\Delta$ \\
\midrule
\multirow{3}{*}{ER}
 & CIFAR-10 & $88.57{\pm}0.92$ & $91.44{\pm}0.04$ & $+2.9$ \\
 & CIFAR-100 & $20.90{\pm}0.84$ & $35.27{\pm}0.19$ & $+14.4$ \\
 & Tiny-IN & $13.53{\pm}0.57$ & $16.88{\pm}0.27$ & $+3.3$ \\
\midrule
\multirow{3}{*}{Uniform}
 & CIFAR-10 & $59.13{\pm}2.86$ & $62.11{\pm}3.12$ & $+3.0$ \\
 & CIFAR-100 & $7.64{\pm}0.82$ & $16.82{\pm}1.04$ & $+9.2$ \\
 & Tiny-IN & $4.56{\pm}0.46$ & $9.74{\pm}0.37$ & $+5.2$ \\
\bottomrule
\end{tabular}
\end{table}

Table~\ref{tab:mlp} reports the extreme point $99.9\%$: relative to the
unguided Magnitude/Random baseline, reachability-guided updates improve
top-5 accuracy by $+3.0$ to $+9.2$ points under uniform initialization and by $+2.9$ to $+14.4$ under ER. The CIFAR-100 ER sparsity sweep (Appendix~\ref{app:results}) shows that the two methods are nearly tied at $99\%$, while the gap opens from $99.75\%$ onward and peaks at $99.9\%$ ($35.27\%$ vs.\ $20.90\%$)---exactly where functional reachability collapses in Figure~\ref{fig:reachability-dst}, consistent with the static cliff in Figure~\ref{fig:reachability}. The same pattern holds on CIFAR-10 and Tiny ImageNet under both initializations (Appendix~\ref{app:results}).
\begin{table}[h]
\centering
\scriptsize
\setlength{\tabcolsep}{3pt}
\renewcommand{\arraystretch}{0.95}
\caption{Top-3 test accuracy ($\%$, mean$\pm$std over five seeds) of an
MLP-Mixer trained on CIFAR-100 under ER initialization: dense baseline,
Magnitude/Gradient, Magnitude/Random, and reachability-guided
Magnitude/Random.}\label{tab:mixer}
\begin{tabular}{lcccc}
\toprule
Sparsity & Dense & Mag./Grad. & Mag./Rand. & G-Mag./Rand. \\
\midrule
$0\%$ & $66.33{\pm}1.11$ & -- & -- & -- \\
$99.00\%$ & -- & $71.92{\pm}1.70$ & $68.73{\pm}0.85$ & $68.12{\pm}0.45$ \\
$99.90\%$ & -- & $61.08{\pm}0.81$ & $45.81{\pm}0.72$ & $57.48{\pm}0.86$ \\
$99.95\%$ & -- & $42.31{\pm}2.89$ & $31.96{\pm}1.11$ & ${47.65{\pm}0.76}$ \\
$99.99\%$ & -- & $13.28{\pm}3.59$ & $4.22{\pm}1.21$ & ${13.56{\pm}1.70}$ \\
\bottomrule
\end{tabular}
\end{table}

Table~\ref{tab:mixer} reports top-3 accuracy.
Gradient regrowth leads at $99\%$--$99.9\%$ sparsity, but from $99.95\%$ onward guided DST overtakes both unguided baselines and remains competitive with gradient regrowth even at $99.99\%$, where random regrowth degrades sharply.

\section{Conclusion}
\label{sec:disc}

We studied extreme-sparse dynamic training through the lens of functional
reachability, which captures whether active connections participate in complete
input--output paths. The resulting mean-field analysis closely matches the
empirical reachability measurements and identifies the sparsity regime in which
unguided DST loses functional connectivity. Using this structure to guide removal and
regrowth preserves substantially more functional pathways and improves accuracy
across MLP experiments, with gains becoming largest at extreme sparsity and
transferring to MLP-Mixer up to $99.99\%$ sparsity. These results suggest that,
when connectivity becomes the limiting resource, DST benefits from explicitly
accounting for the global structural role of its active edges. Extending this
notion to architectures with residual connections remains an important
direction for future work.

\bibliography{refs}

\newpage
\appendix
\input{appendix}

\end{document}

%% file: commenting_macros.tex
\newcommand{\hide}[1]{}

\usepackage{xcolor}
\usepackage{amssymb}
\usepackage{wasysym}

\definecolor{mohammadcolor}{rgb}{0.0,0.0,1.0}
\definecolor{jeroencolor}{rgb}{0.85,0.6,0.0}
\definecolor{mateuszcolor}{rgb}{0.722, 0.11, 0.42}
\definecolor{elenacolor}{rgb}{0.722, 0.11, 0.42}
\definecolor{todocolor}{rgb}{0.9,0.1,0.1}
\definecolor{changedcolor}{rgb}{0.2,0.2,1}

\newcommand{\redacted}[1]{\emph{[anonymized for review]}}



%% file: appendix.tex

\section{Reachability formalism}
\label{app:theory_formalism}

This section gives the full definitions and derivations summarized in
Section~\ref{sec:theory}.
We model a feed-forward neural network as a layered graph $G=(V,E)$ with
$V=\bigcup_{l=0}^{L}V_l$, candidate edges
$\mathcal{U}^{e}=\bigcup_{l=1}^{L}(V_{l-1}\times V_l)$, and $E\subseteq\mathcal{U}^{e}$.
Edges between $V_{l-1}$ and $V_l$ appear independently with probability
$p_l\in[0,1]$ (covering uniform and Erd\H{o}s--R\'enyi / ER densities).

\paragraph{Forward, backward, and functional sets.}
Forward- and backward-reachable vertices are defined recursively from the input and output layers:
\begin{align*}
R^{\mathrm{fwd},v}_{0}&=V_0,&
R^{\mathrm{fwd},v}_{l}
&=\bigl\{v\in V_l \mid \exists\,u\in R^{\mathrm{fwd},v}_{l-1}:\,(u,v)\in E\bigr\},\\
R^{\mathrm{bwd},v}_{L}&=V_L,&
R^{\mathrm{bwd},v}_{l}
&=\bigl\{u\in V_l \mid \exists\,v\in R^{\mathrm{bwd},v}_{l+1}:\,(u,v)\in E\bigr\}.
\end{align*}
Functional vertices are their intersection,
$R^{\mathrm{fcnl},v}_{l}=R^{\mathrm{fwd},v}_{l}\cap R^{\mathrm{bwd},v}_{l}$,
i.e., neurons that lie on at least one input--output path and can therefore
participate in both a forward pass and the corresponding backward signal.
Endpoint-consistent edge candidates and active reachable edges are
\begin{align*}
\widetilde{R}^{\mathrm{fwd},e}_{l}
&=\{(u,v)\in V_{l-1}\times V_l
\mid u\in R^{\mathrm{fwd},v}_{l-1},\; v\in R^{\mathrm{fwd},v}_{l}\},\\
\widetilde{R}^{\mathrm{bwd},e}_{l}
&=\{(u,v)\in V_{l-1}\times V_l
\mid u\in R^{\mathrm{bwd},v}_{l-1},\; v\in R^{\mathrm{bwd},v}_{l}\},
\end{align*}
with
$R^{\mathrm{fwd},e}_{l}=\widetilde{R}^{\mathrm{fwd},e}_{l}\cap E$
and
$R^{\mathrm{bwd},e}_{l}=\widetilde{R}^{\mathrm{bwd},e}_{l}\cap E$.
Functional candidates and active functional edges follow as
\begin{equation*}
\widetilde{R}^{\mathrm{fcnl},e}_{l}
=\widetilde{R}^{\mathrm{fwd},e}_{l}\cap\widetilde{R}^{\mathrm{bwd},e}_{l},\qquad
R^{\mathrm{fcnl},e}_{l}
=\widetilde{R}^{\mathrm{fcnl},e}_{l}\cap E
=R^{\mathrm{fwd},e}_{l}\cap R^{\mathrm{bwd},e}_{l}.
\end{equation*}

\paragraph{Expectations.}
To quantify how these sets shrink under random sparsity, we next derive
expectations under independent Bernoulli edges.
Let $F_{l-1}=|R^{\mathrm{fwd},v}_{l-1}|$.
Conditioning on $F_{l-1}=r$ and using independent Bernoulli edges gives
$\mathbb{E}[|R^{\mathrm{fwd},e}_{l}|\,|\,F_{l-1}=r]=p_l|V_l|r$
and
$\mathbb{E}[|R^{\mathrm{fwd},v}_{l}|\,|\,F_{l-1}=r]=|V_l|\bigl(1-(1-p_l)^{r}\bigr)$.
Unconditioning yields the exact identities
\begin{align}
\mathbb{E}\!\bigl[|R^{\mathrm{fwd},e}_{l}|\bigr]
&=p_l\,|V_l|\,\mathbb{E}\!\bigl[|R^{\mathrm{fwd},v}_{l-1}|\bigr],
\label{eq:E-fwd-e}\\
\mathbb{E}\!\bigl[|R^{\mathrm{fwd},v}_{l}|\bigr]
&=|V_l|\,
\mathbb{E}\!\Bigl[1-(1-p_l)^{|R^{\mathrm{fwd},v}_{l-1}|}\Bigr],
\label{eq:E-fwd-v-exact}
\end{align}
and the mean-field closure
\begin{equation}
\mathbb{E}\!\bigl[|R^{\mathrm{fwd},v}_{l}|\bigr]
\approx
|V_l|\Bigl(1-(1-p_l)^{\mathbb{E}[|R^{\mathrm{fwd},v}_{l-1}|]}\Bigr).
\label{eq:E-fwd-v-mf}
\end{equation}
By symmetry from the output side,
\begin{align}
\mathbb{E}\!\bigl[|R^{\mathrm{bwd},e}_{l}|\bigr]
&=p_l\,|V_{l-1}|\,\mathbb{E}\!\bigl[|R^{\mathrm{bwd},v}_{l}|\bigr],
\label{eq:E-bwd-e}\\
\mathbb{E}\!\bigl[|R^{\mathrm{bwd},v}_{l}|\bigr]
&\approx
|V_l|\Bigl(1-(1-p_{l+1})^{\mathbb{E}[|R^{\mathrm{bwd},v}_{l+1}|]}\Bigr).
\label{eq:E-bwd-v-mf}
\end{align}
Under an independence approximation between the two directions, this yields
\begin{equation}
\mathbb{E}\!\bigl[|R^{\mathrm{fcnl},e}_{l}|\bigr]
\approx
p_l\,\mathbb{E}\!\bigl[|R^{\mathrm{fwd},v}_{l-1}|\bigr]\,
\mathbb{E}\!\bigl[|R^{\mathrm{bwd},v}_{l}|\bigr],
\qquad
\mathbb{E}\!\bigl[|R^{\mathrm{fcnl},v}_{l}|\bigr]
\approx
\frac{\mathbb{E}[|R^{\mathrm{fwd},v}_{l}|]\,\mathbb{E}[|R^{\mathrm{bwd},v}_{l}|]}{|V_l|},
\end{equation}
i.e., \eqref{eq:E-fcnl-e} and~\eqref{eq:E-fcnl-v}.
Thus functional connectivity is limited by the \emph{overlap} of forward and
backward reachability: once either direction thins, the functional fraction
falls multiplicatively rather than additively, predicting sharp connectivity
cliffs at extreme sparsity.

\section{Guided regrowth details}
\label{app:guided_regrowth}

Section~\ref{sec:method} summarizes guided regrowth; here we give the fallback
candidate sets and sampling distribution used when
$R^{\mathrm{guide},e}_l=\varnothing$.
\begin{equation}
\begin{aligned}
R^{\mathrm{fwd\text{-}grow},e}_l
&=
\left\{
(u,v)\in\mathcal{I}_l
\;\middle|\;
u\in R^{\mathrm{fwd},v}_{l-1}(E^{-})
\right\},\;
R^{\mathrm{bwd\text{-}grow},e}_l
&=
\left\{
(u,v)\in\mathcal{I}_l
\;\middle|\;
v\in R^{\mathrm{bwd},v}_{l}(E^{-})
\right\}.
\end{aligned}
\label{eq:fallback-sets}
\end{equation}
The effective guided candidate set is
\begin{equation}
\widetilde{R}^{\mathrm{guide},e}_l
=
\begin{cases}
R^{\mathrm{guide},e}_l,
& \text{if } R^{\mathrm{guide},e}_l\neq\varnothing,\\[2mm]
R^{\mathrm{fwd\text{-}grow},e}_l
\cup
R^{\mathrm{bwd\text{-}grow},e}_l,
& \text{otherwise}.
\end{cases}
\label{eq:effective-guide-set}
\end{equation}
This fallback propagates connectivity forward from the input and backward
from the output, allowing the two reachable regions to eventually meet and
restore functional input--output connectivity.
The remaining inactive edges are
$R^{\mathrm{ung},e}_l=\mathcal{I}_l\setminus\widetilde{R}^{\mathrm{guide},e}_l$.
With guidance strength $\beta\in[0,1]$ (we use $\beta=0.99$), we sample the
$k_l$ replacement edges from the normalized distribution with unnormalized
masses
\begin{equation}
\tilde{p}^{(l)}_{uv}
=
\begin{cases}
\beta\big/|\widetilde{R}^{\mathrm{guide},e}_{l}|
& (u,v)\in \widetilde{R}^{\mathrm{guide},e}_{l},\\[2pt]
(1-\beta)\big/|R^{\mathrm{ung},e}_{l}|
& (u,v)\in R^{\mathrm{ung},e}_{l}.
\end{cases}
\label{eq:regrow-p}
\end{equation}

\section{Experimental details}
\label{app:setup}

\paragraph{Overview.}
The main text compares reachability-guided DST with an unguided
Magnitude/Random baseline on $4$-layer MLPs trained on CIFAR-10, CIFAR-100,
and Tiny ImageNet, using both ER and uniform mask initialization at sparsities
ranging from $99\%$ to $99.9\%$.
We further evaluate transfer to an MLP-Mixer on CIFAR-100 under ER
initialization, up to $99.99\%$ sparsity.
For the MLP experiments we also report static sparsity (fixed mask) and
parameter-matched dense baselines.
Extended accuracy tables and the CIFAR-100 ER top-5 sparsity sweep appear inAppendix~\ref{app:results}.

\paragraph{Hardware.}
We have used a Dell PowerEdge R750 with 72 CPU cores (144 threads), 256\,GB RAM,
and two NVIDIA A16 GPUs for all the runs.

\paragraph{MLP architecture.}
Sparse models are four-layer MLPs with dimensions
$[I,1024]$, $[1024,512]$, $[512,512]$, $[512,O]$, where $I$ is the flattened
input size and $O\in\{10,100,200\}$ for CIFAR-10, CIFAR-100, and Tiny ImageNet.
ReLU and LayerNorm follow each hidden linear layer; the output layer is dense
linear.
The parameter-matched dense baseline is a three-layer MLP whose parameter count
is at least that of the sparse model at the target sparsity
(following~\cite{nowak2023fantastic}).

\paragraph{MLP training protocol.}
we have used cross-entropy as the runs loss function; SGD with learning rate $0.01$, momentum $0$,
weight decay $5\times 10^{-4}$; exponential LR decay $\gamma=0.985$; batch size
$128$; $100$ epochs; prune fraction $p=0.25$ every $\Delta t=800$ mini-batches;
guidance strength $\beta=0.99$.
Masks are initialized with ER or uniform densities at sparsity
$\{99,99.5,99.75,99.875,99.9\}\%$.
Unguided DST uses magnitude removal and random regrowth; guided DST uses the
operators in Section~\ref{sec:method}.
We also report static sparsity (fixed mask) for reference.
Accuracy/loss are mean$\pm$std over multiple seeds.

\paragraph{MLP-Mixer.}
On CIFAR-100 we use an MLP-Mixer with patch size $4$, $10$ mixer blocks,
token/channel mixing width $512$, and MLP expansion $2048$, with an extra
LayerNorm in the channel-mixing MLP for sparse stability.
Optimization: SGD (lr $0.007$, momentum $0.9$, weight decay $10^{-4}$),
multi-step LR decay at epochs $\{125,187\}$ with factor $0.5$, batch $128$,
$250$ epochs, ER initialization.
Topology updates use $\Delta t=1500$ for magnitude variants and $\Delta t=3000$
for gradient regrowth; prune fraction $0.25$.
A pre-training topology warmup of $\{0,1,5,10,15\}$ epochs is applied with
longer warmup at higher sparsity.
Reachability guidance is computed on each active bipartite projection block
using the same forward/backward definitions as in the MLP case.
Methods compared: Magnitude/Gradient, Magnitude/Random, and
Guided-Magnitude/Random ($\beta=0.99$), each over five runs.

\paragraph{Monte Carlo validation.}
For Figure~\ref{fig:reachability} and Figures~\ref{fig:theory-er}--\ref{fig:theory-uniform},
sparsity is swept over $[0.99,0.9999]$ on $100$ uniform grid points.
At each point we draw $100$ random masks on the experimental MLP widths
$[3072,1024,512,512,100]$,
compute forward/backward/functional edge fractions per layer, and plot
mean$\pm$std (``realized'') against the mean-field predictions from
Appendix~\ref{app:theory_formalism} evaluated at the realized per-layer densities
(``predicted'').

\section{Monte Carlo reachability grids}
\label{app:theory_mc}

\begin{figure}[ht]
\centering
\setlength{\tabcolsep}{1.5pt}%
\renewcommand{\arraystretch}{0.9}%
\begin{tabular}{@{}ccc@{}}
\includegraphics[width=0.32\linewidth]{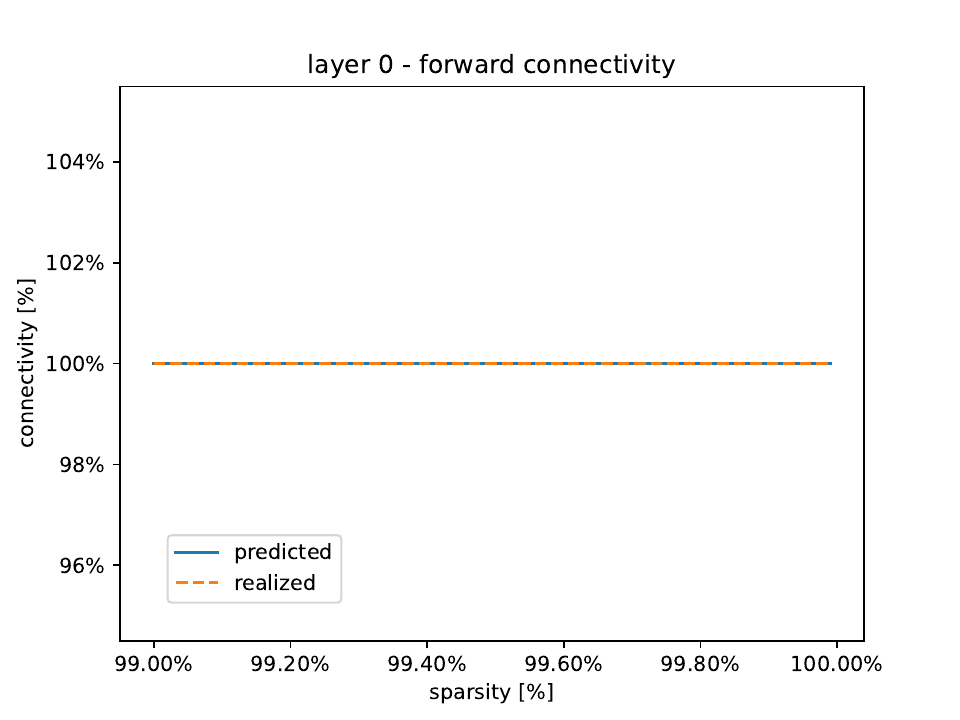} &
\includegraphics[width=0.32\linewidth]{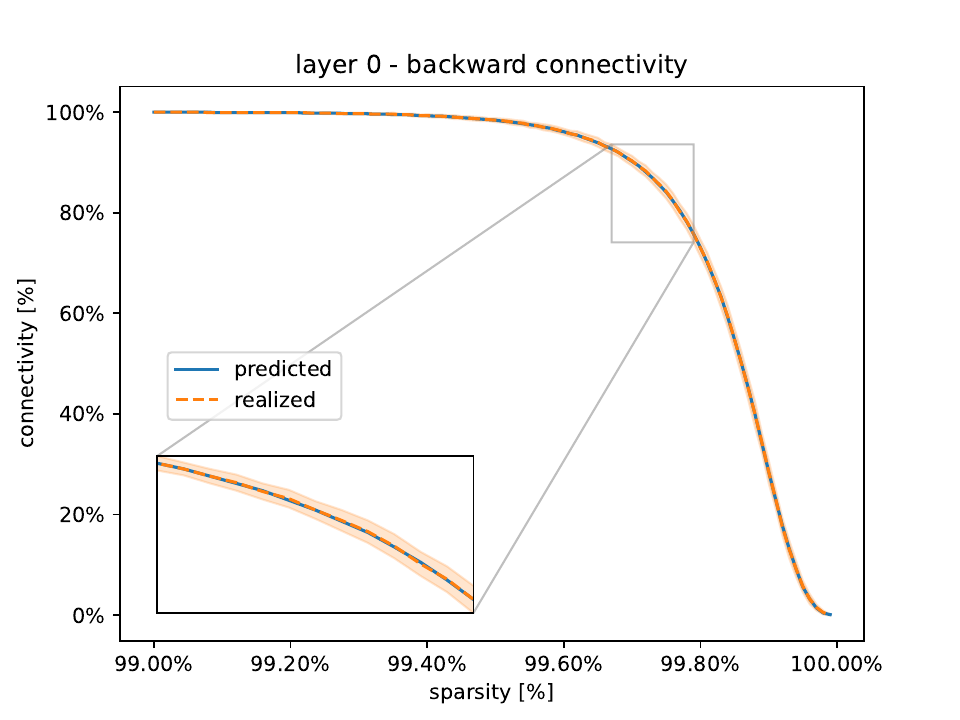} &
\includegraphics[width=0.32\linewidth]{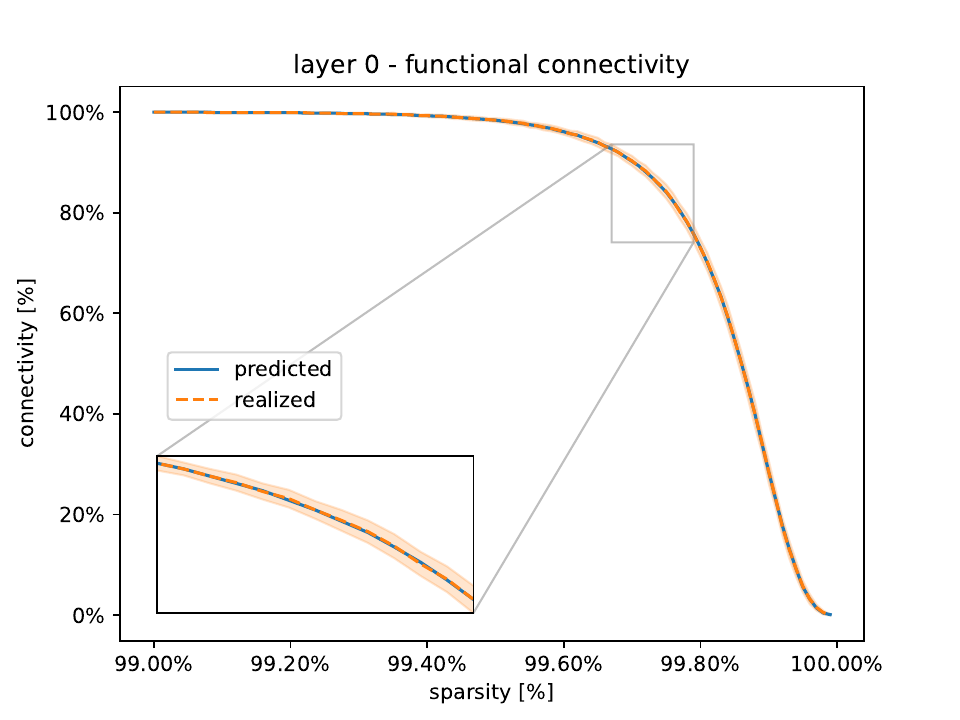} \\
\includegraphics[width=0.32\linewidth]{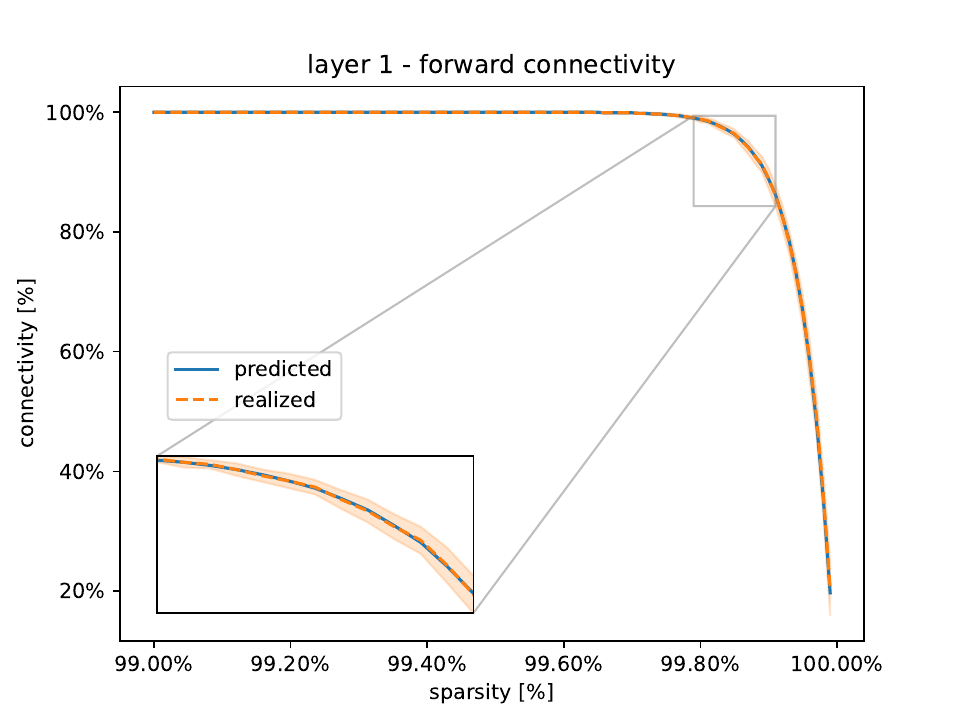} &
\includegraphics[width=0.32\linewidth]{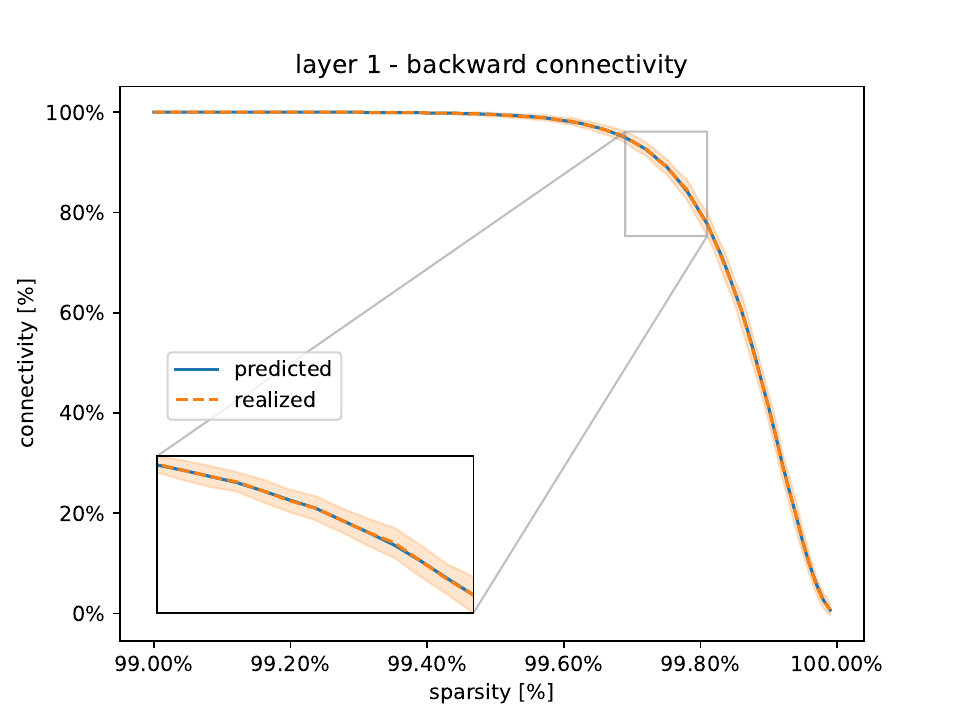} &
\includegraphics[width=0.32\linewidth]{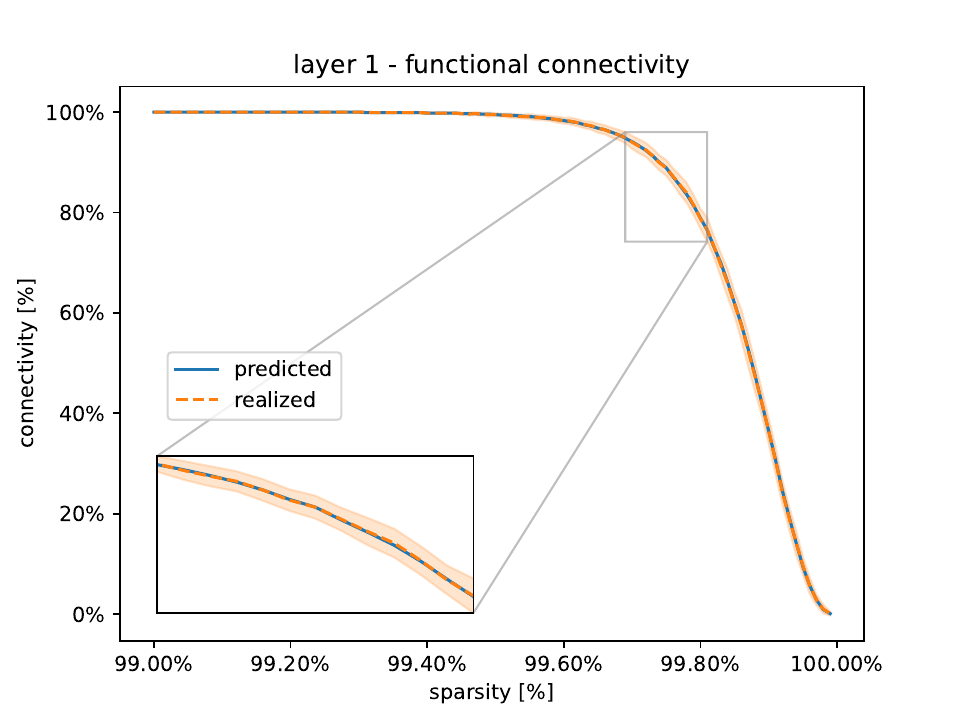} \\
\includegraphics[width=0.32\linewidth]{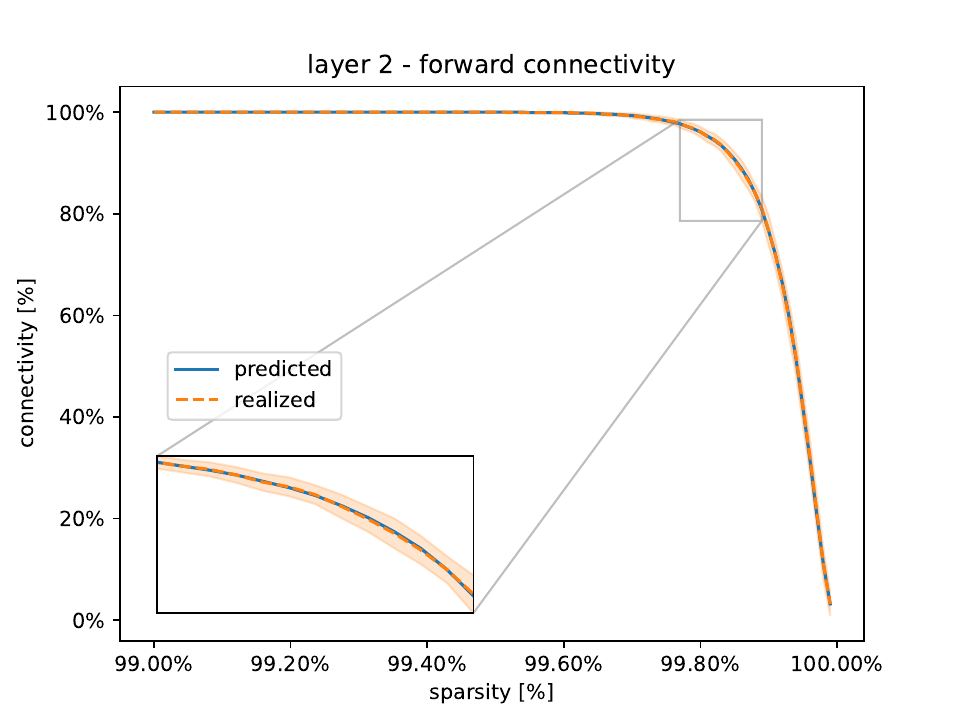} &
\includegraphics[width=0.32\linewidth]{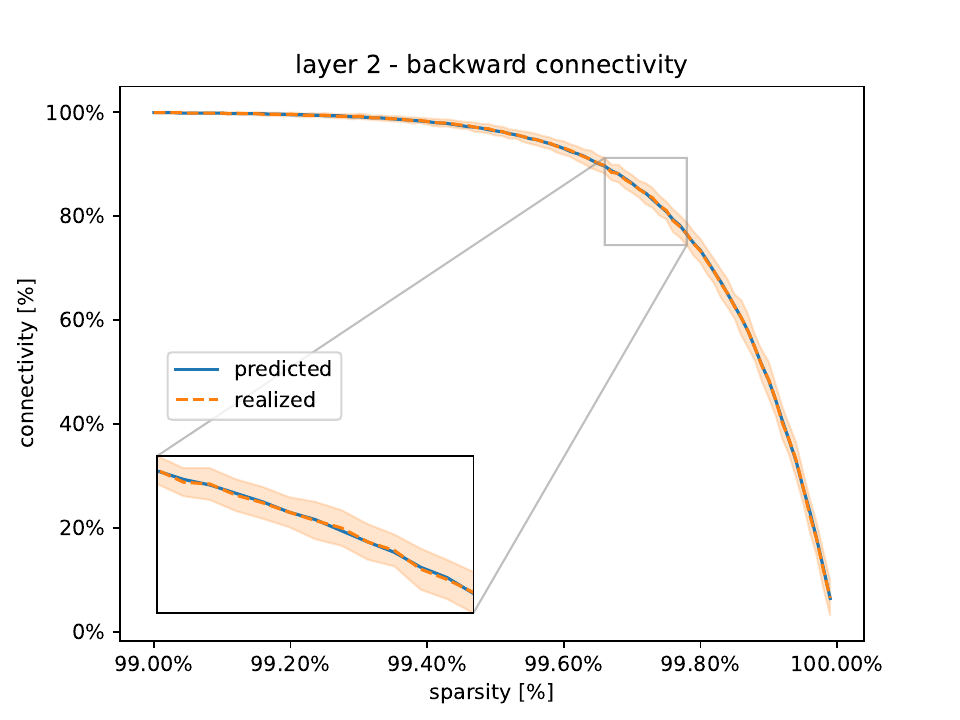} &
\includegraphics[width=0.32\linewidth]{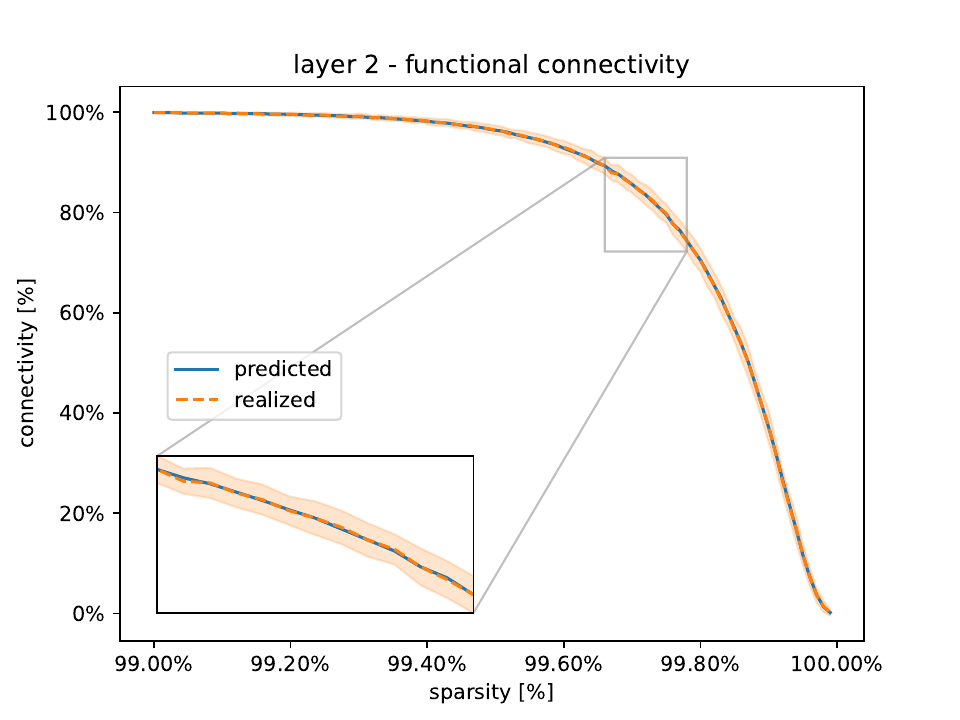} \\
\includegraphics[width=0.32\linewidth]{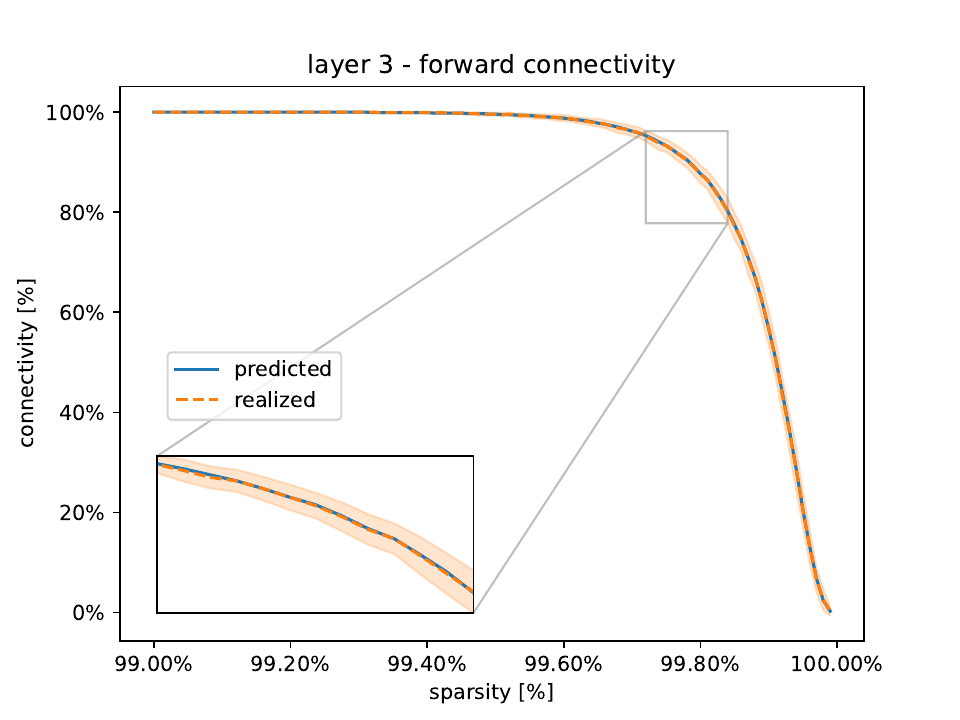} &
\includegraphics[width=0.32\linewidth]{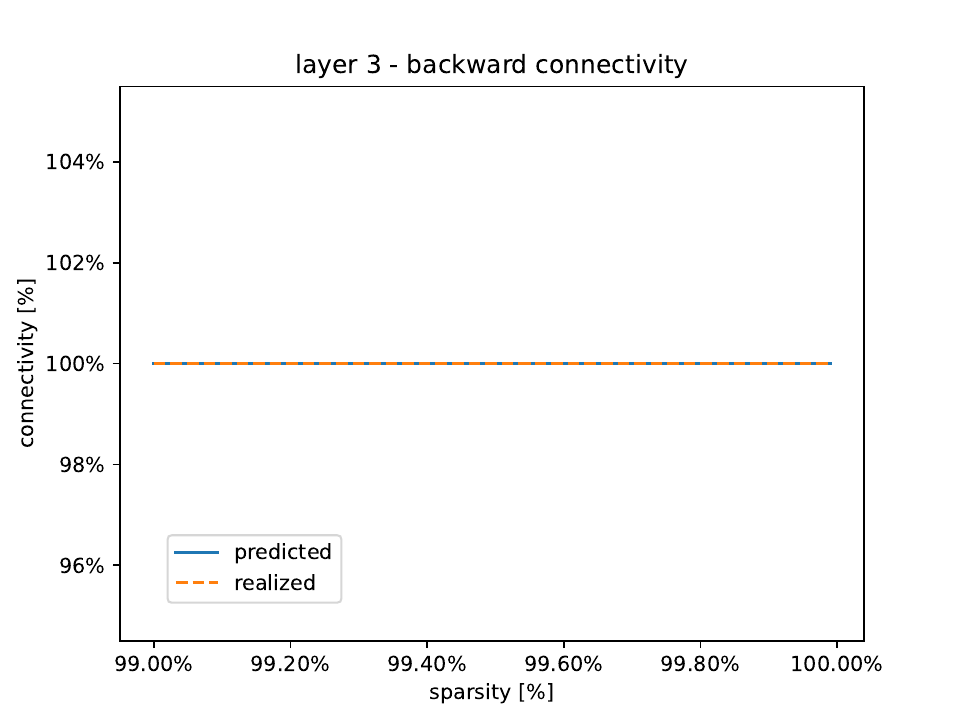} &
\includegraphics[width=0.32\linewidth]{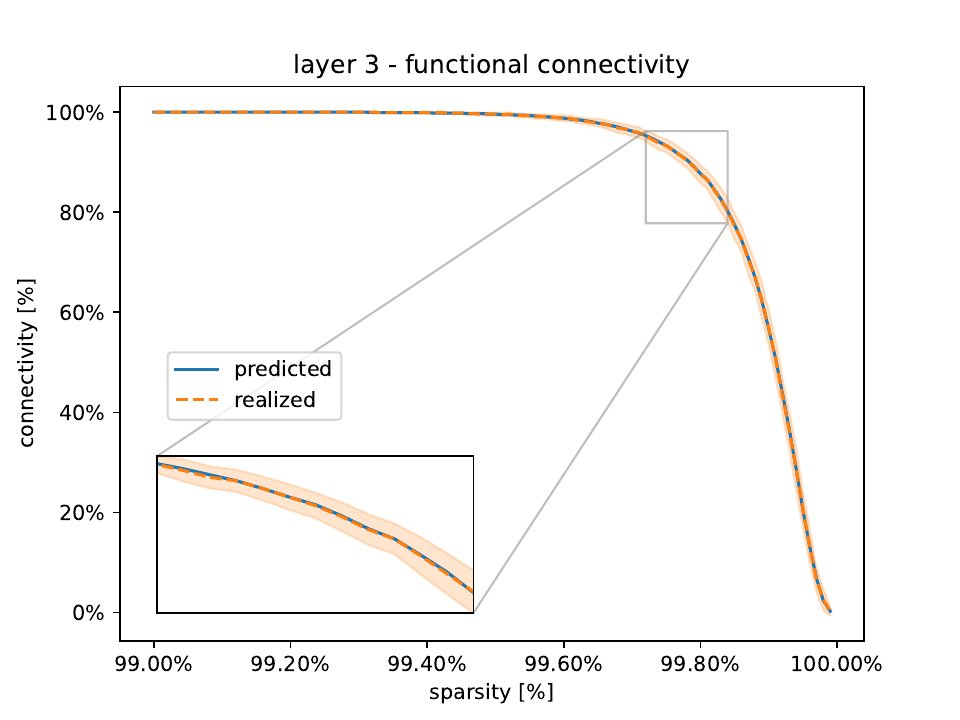}
\end{tabular}
\caption{ER initialization: forward, backward, and functional edge reachability
(rows: layers~$0$--$3$; columns: forward / backward / functional).
Predicted (solid) vs.\ Monte Carlo realized (dashed, $\pm$std).}
\label{fig:theory-er}
\end{figure}

\begin{figure}[ht]
\centering
\setlength{\tabcolsep}{1.5pt}%
\renewcommand{\arraystretch}{0.9}%
\begin{tabular}{@{}ccc@{}}
\includegraphics[width=0.32\linewidth]{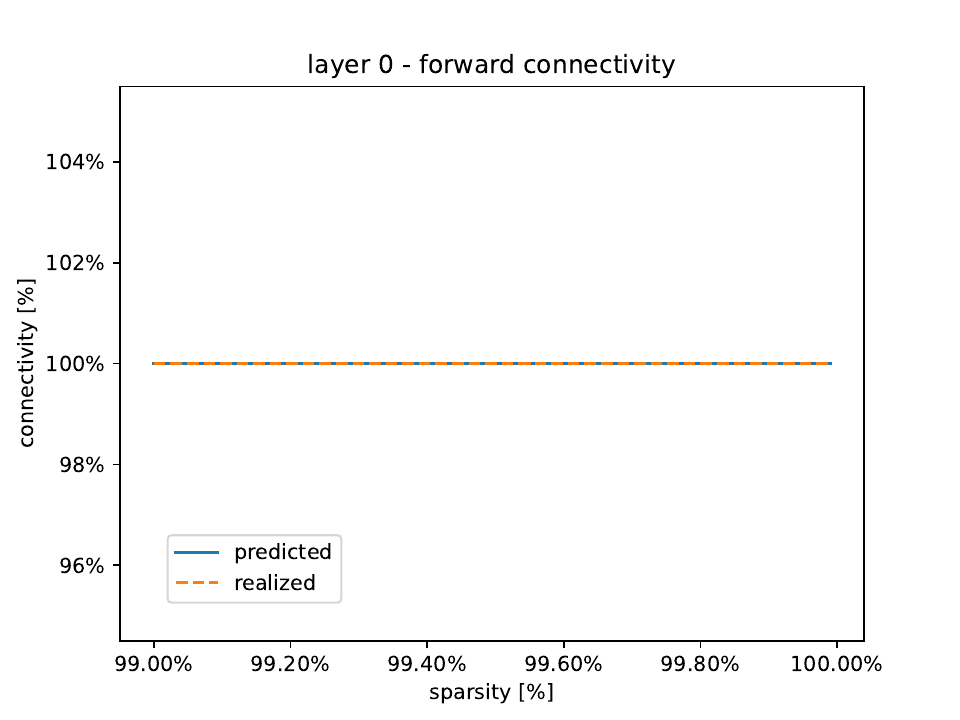} &
\includegraphics[width=0.32\linewidth]{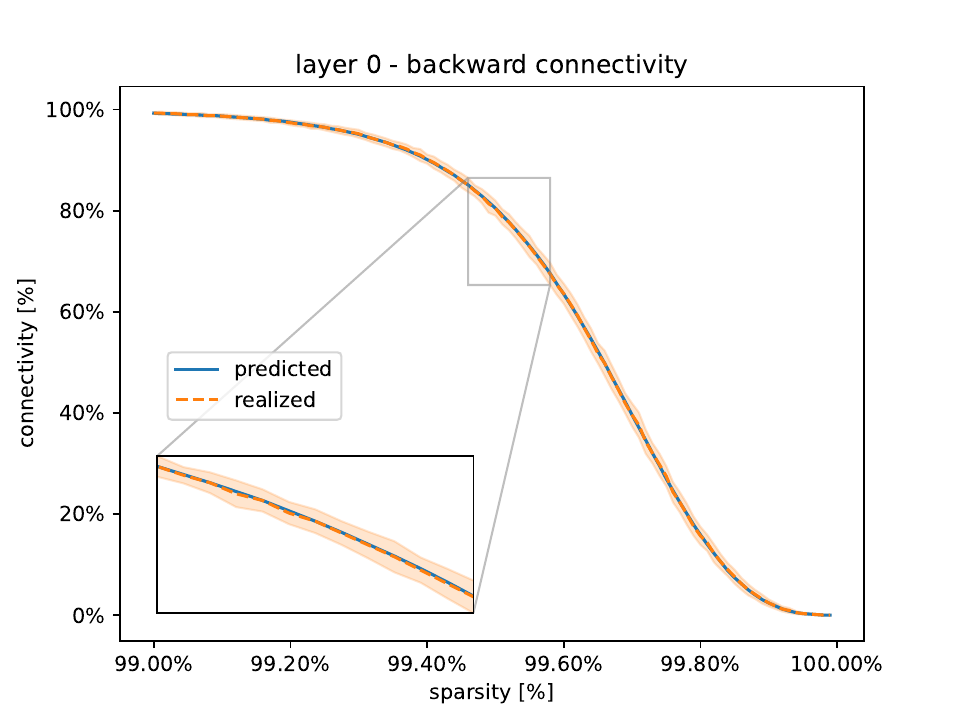} &
\includegraphics[width=0.32\linewidth]{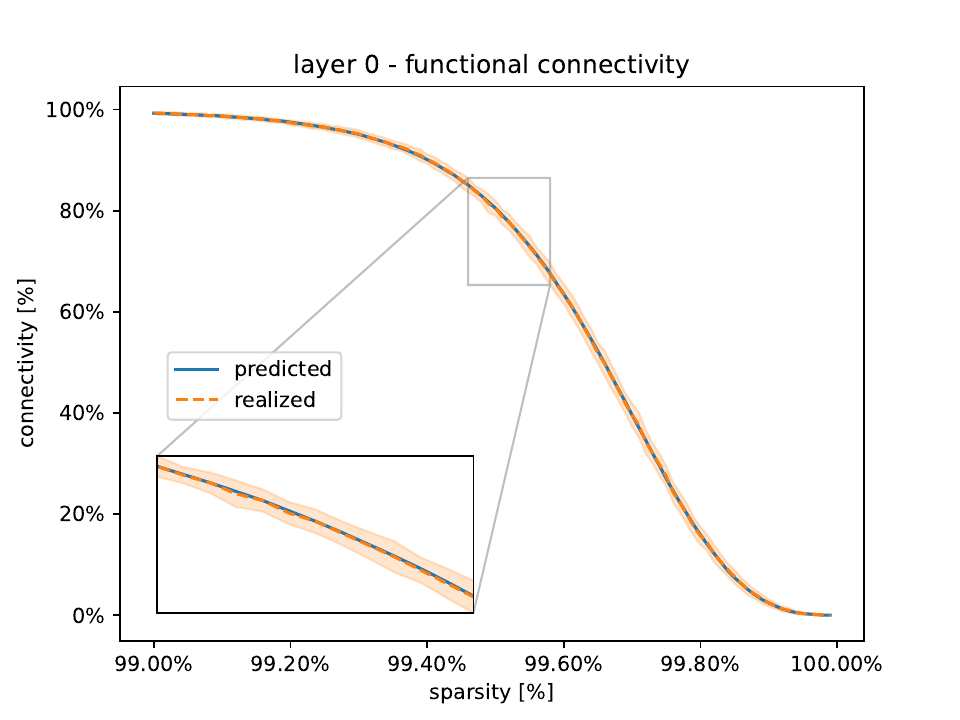} \\
\includegraphics[width=0.32\linewidth]{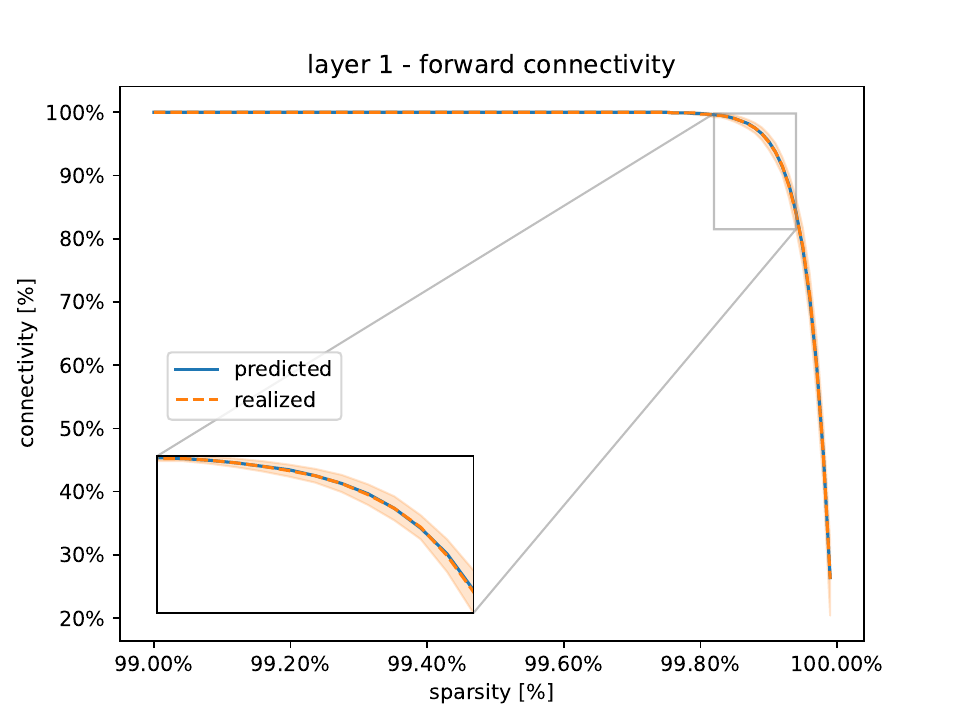} &
\includegraphics[width=0.32\linewidth]{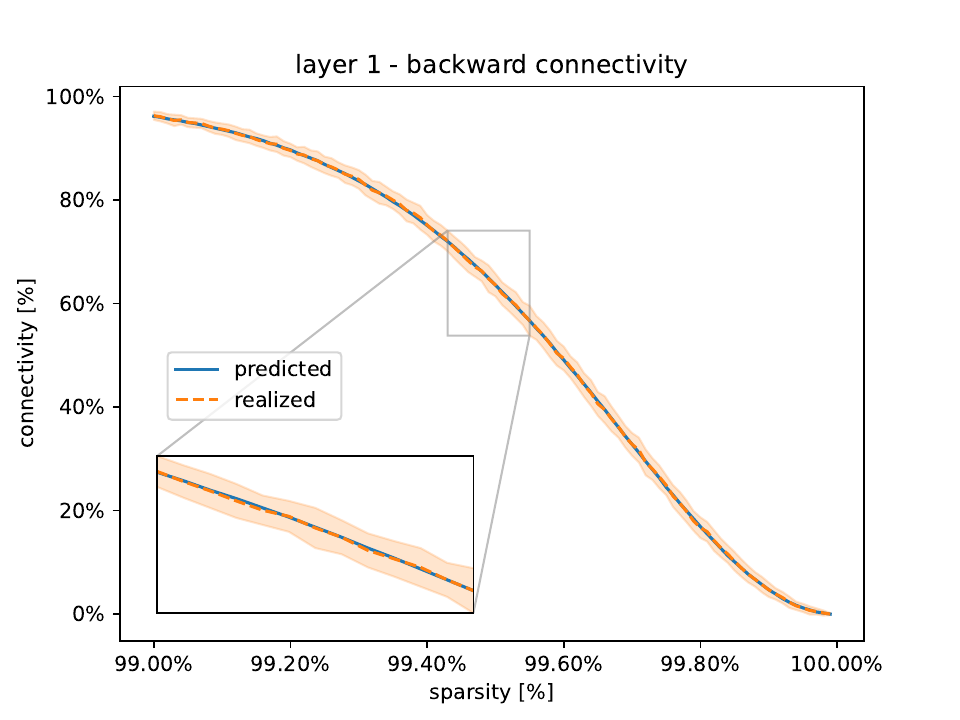} &
\includegraphics[width=0.32\linewidth]{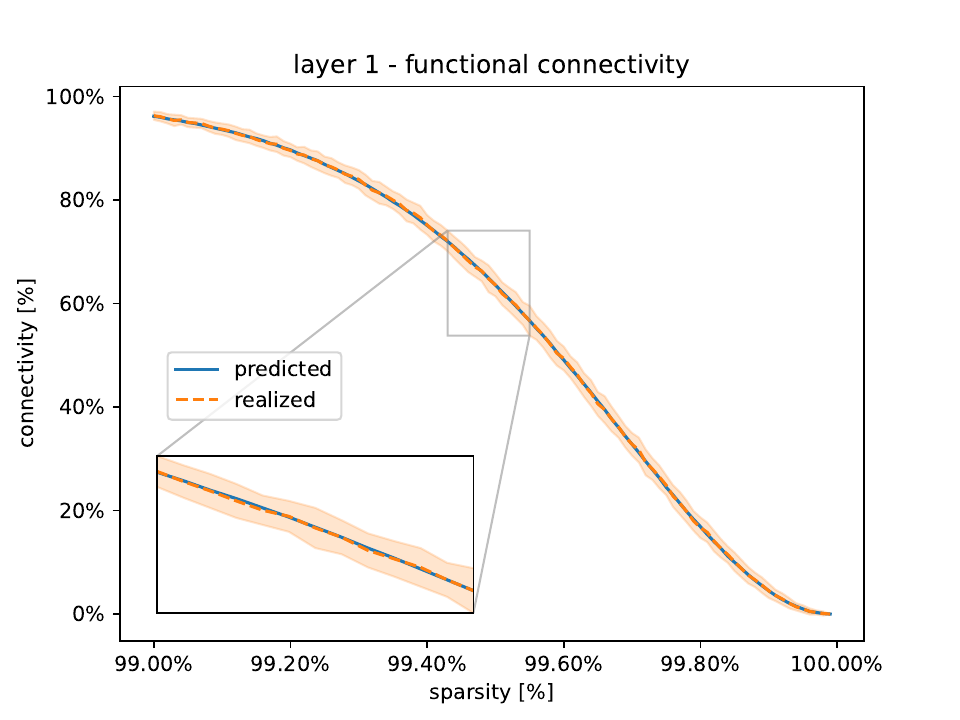} \\
\includegraphics[width=0.32\linewidth]{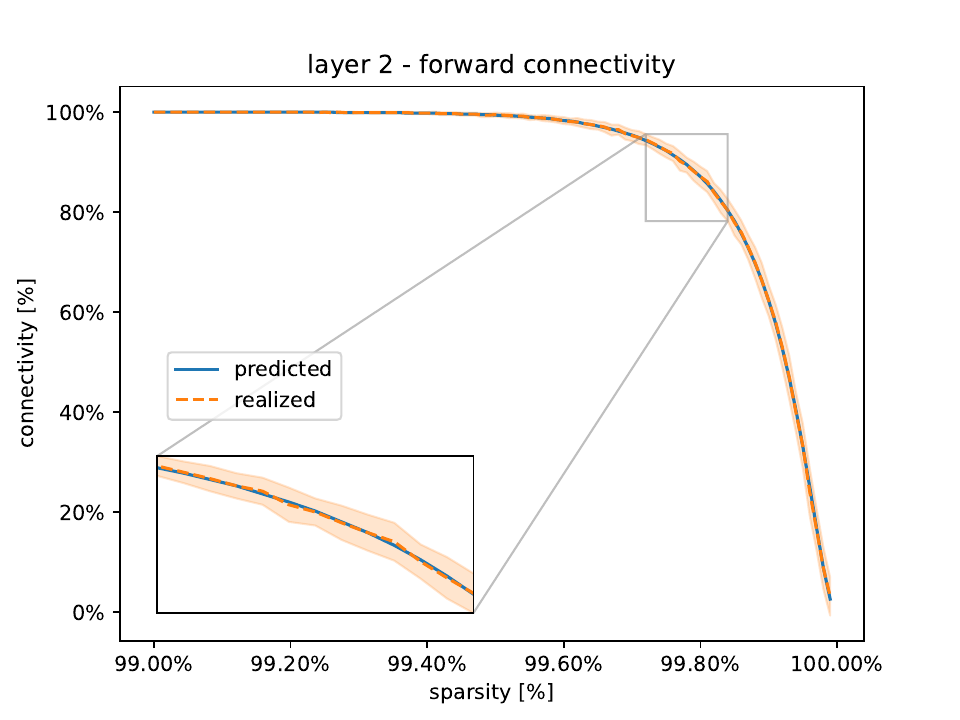} &
\includegraphics[width=0.32\linewidth]{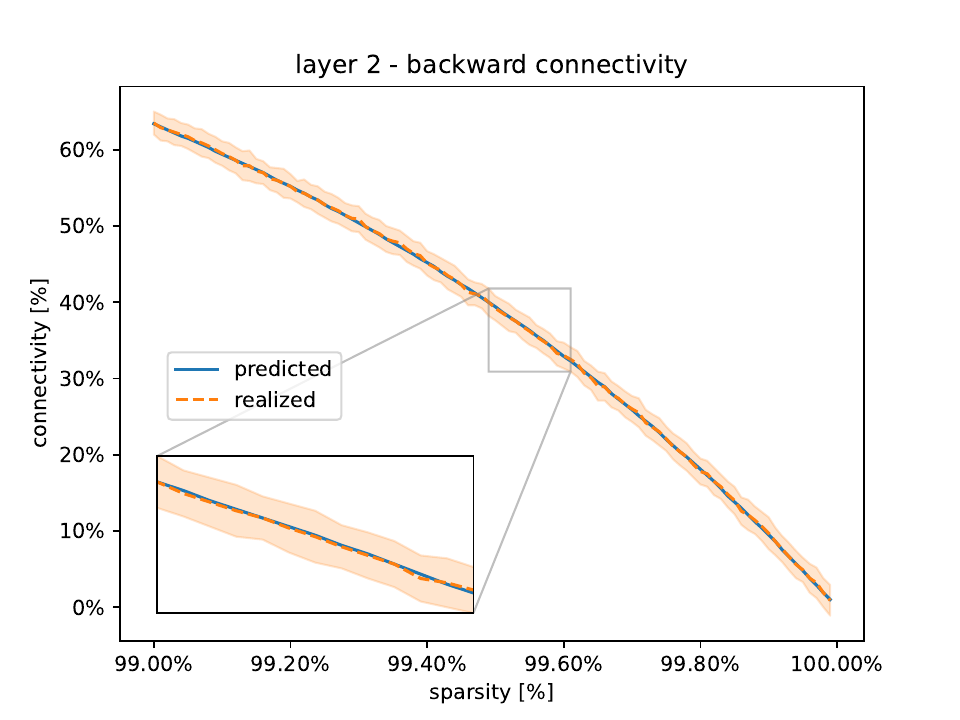} &
\includegraphics[width=0.32\linewidth]{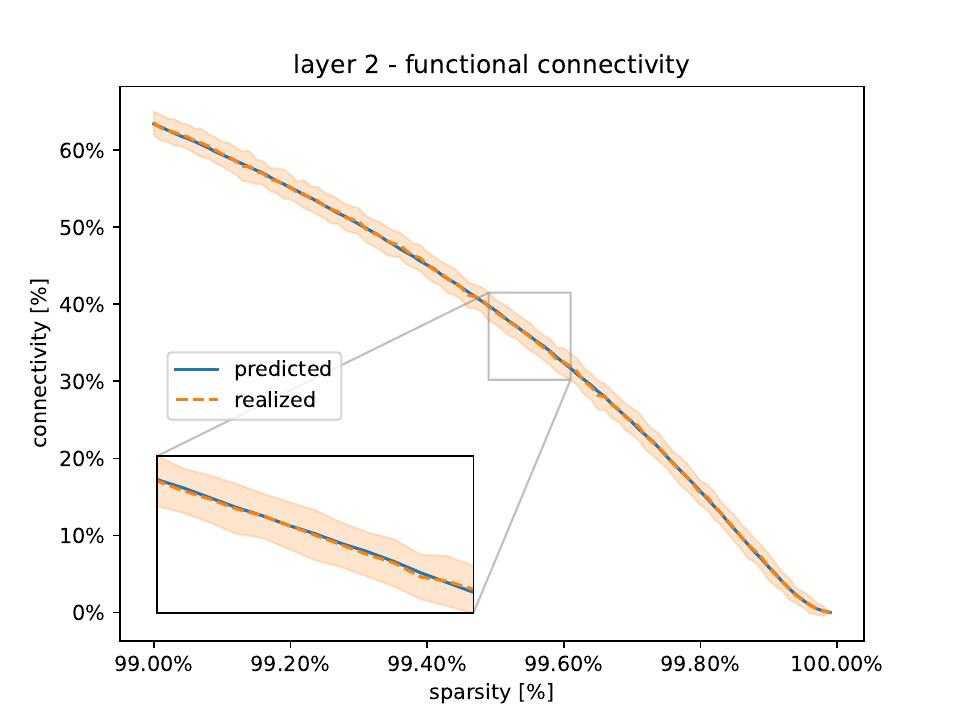} \\
\includegraphics[width=0.32\linewidth]{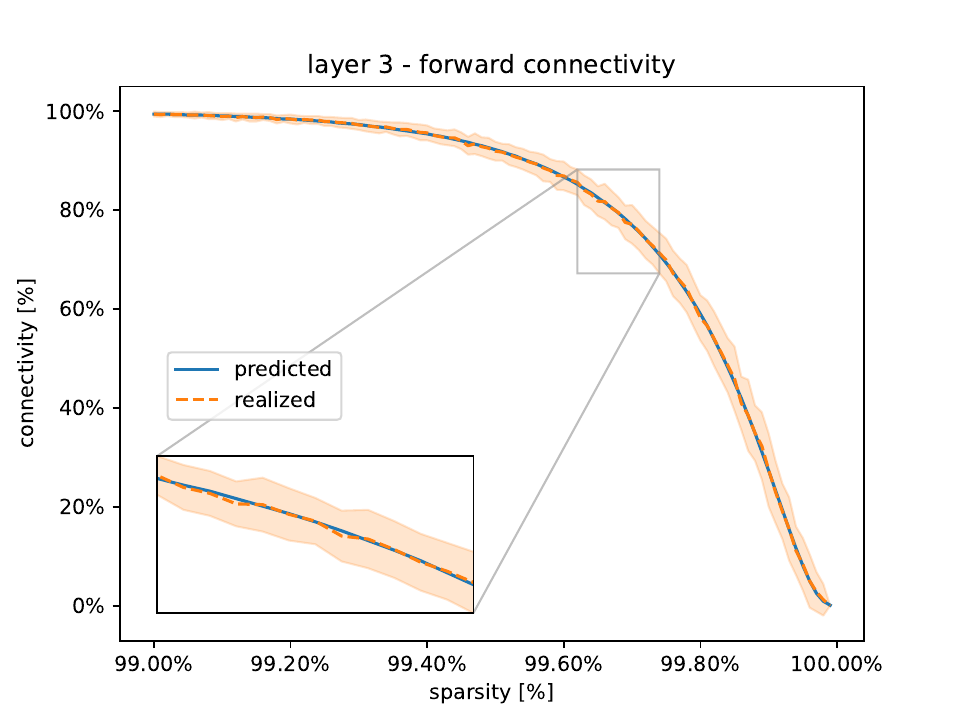} &
\includegraphics[width=0.32\linewidth]{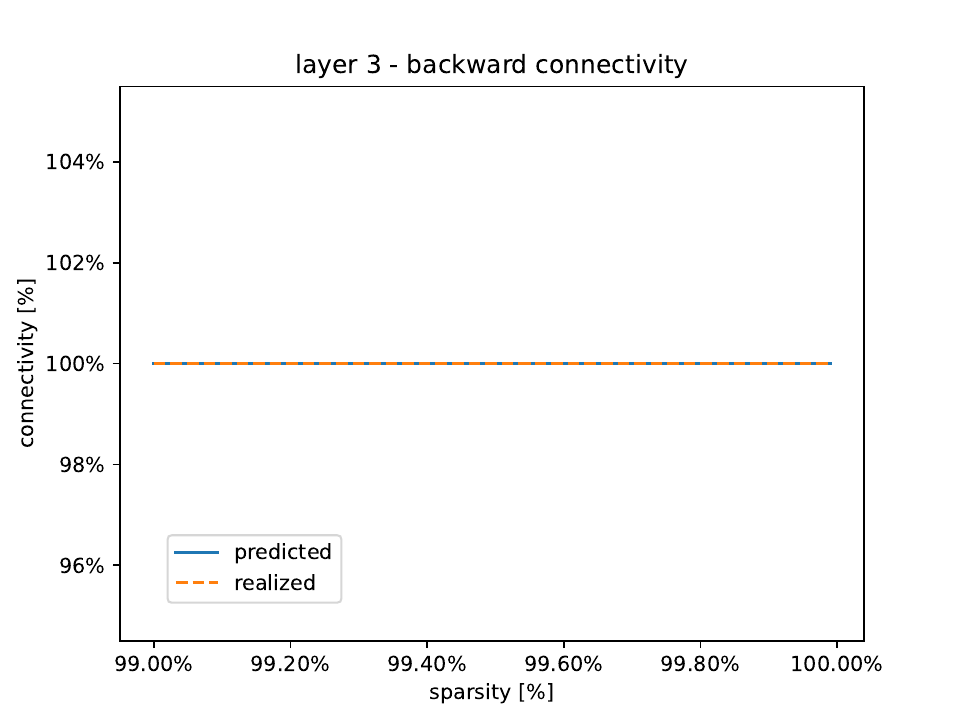} &
\includegraphics[width=0.32\linewidth]{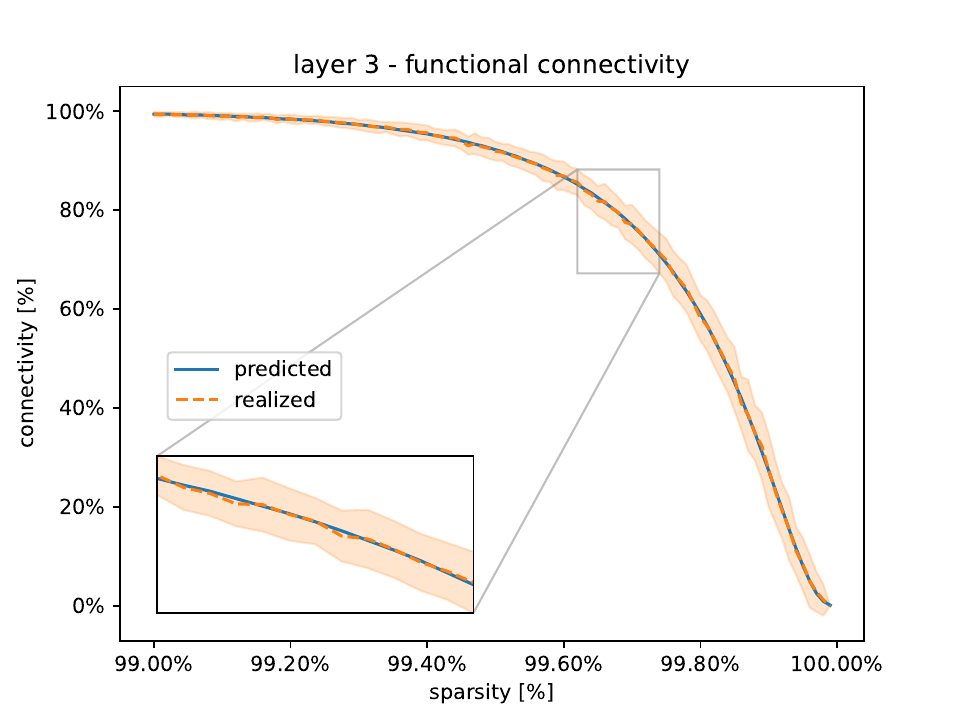}
\end{tabular}
\caption{Uniform initialization: same layout as Figure~\ref{fig:theory-er}.
Functional reachability decays earlier and more gradually than under ER.}
\label{fig:theory-uniform}
\end{figure}

The main text (Figure~\ref{fig:reachability}) shows functional reachability for
ER initialization on layers~$1$--$2$ only.
Figures~\ref{fig:theory-er} and~\ref{fig:theory-uniform} give the complete
forward / backward / functional panels for every interlayer bipartite graph
(layers~$0$--$3$) under ER and uniform initialization.
By construction, forward reachability in layer~$0$ and backward reachability
in layer~$3$ remain trivial ($=1$) whenever the corresponding edges are active.
Functional reachability combines both constraints and exhibits the
initialization-dependent collapse discussed in Section~\ref{sec:theory}.

\section{Empirical reachability during training}
\label{app:reachability}

We compute functional edge fractions directly from training logs: for each
masked layer $l$, we count active edges in
$R^{\mathrm{fwd},e}_l\cap R^{\mathrm{bwd},e}_l$ and divide by the number of
active edges.
Network-level curves average this quantity over all masked layers at the end
of each epoch.

\begin{figure}[h]
\centering
\includegraphics[width=\linewidth]{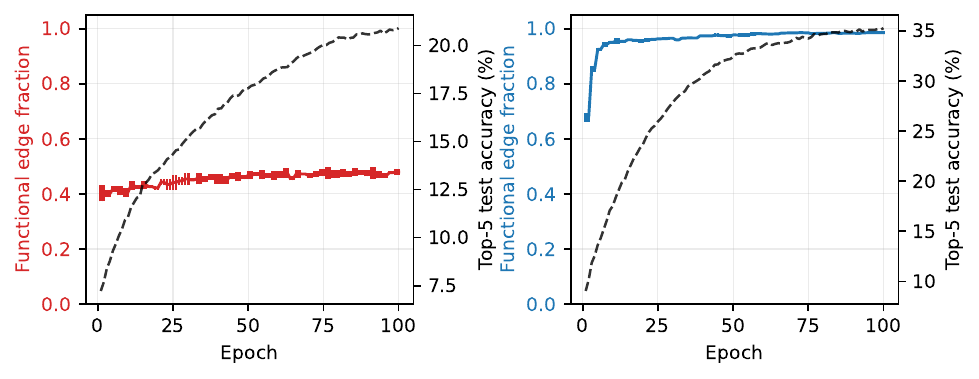}
\caption{MLP CIFAR-100, ER init, $99.9\%$ sparsity: epoch-wise functional edge
fraction (solid) and test accuracy (dashed).}
\label{fig:app-train-999}
\end{figure}

\begin{figure}[h]
\centering
\includegraphics[width=0.72\linewidth]{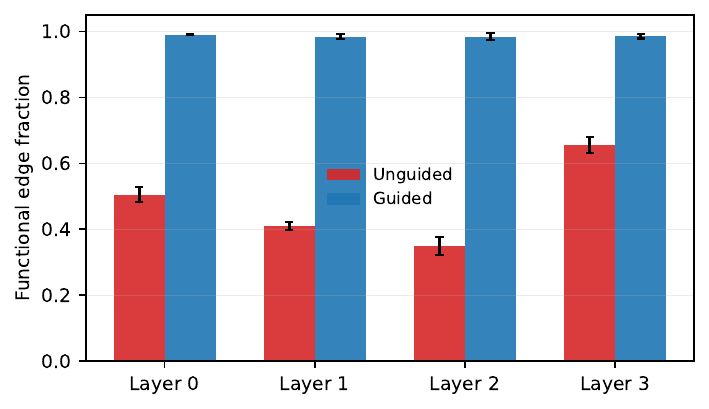}
\caption{Per-layer functional edge fraction at $99.9\%$ sparsity (ER, final
epoch, five seeds).}
\label{fig:app-layers-999}
\end{figure}

\paragraph{Training dynamics at $99.9\%$.}
Figure~\ref{fig:app-train-999} shows epoch-wise functional fractions and
accuracy for ER-init MLPs at $99.9\%$ sparsity.
Unguided DST starts near-perfect but the functional fraction decays as training
progresses; guided DST keeps both metrics high.
Figure~\ref{fig:app-layers-999} decomposes the final epoch by layer: unguided
DST concentrates the loss in middle layers ($\approx\!35$--$50\%$ functional
edges), whereas guided DST keeps every layer above $\approx\!98\%$.

\section{Extended results}
\label{app:results}

Tables~\ref{tab:app-er}--\ref{tab:app-unif} report DST test accuracy across the
full sparsity grid for ER and uniform initialization.
Guided and unguided methods are comparable near $99\%$; guided DST pulls ahead
as sparsity increases, with the largest absolute gains at $99.875\%$--$99.9\%$,
consistent with the functional-reachability cliff in Figure~\ref{fig:reachability}.

\begin{table}[ht]
\centering
\scriptsize
\caption{DST top-1 accuracy ($\%$) under \textbf{ER} initialization:
Magnitude/Random (U) vs.\ reachability-guided (G).}
\label{tab:app-er}
\begin{tabular}{llccccc}
\toprule
Dataset & & $99\%$ & $99.5\%$ & $99.75\%$ & $99.875\%$ & $99.9\%$ \\
\midrule
\multirow{2}{*}{CIFAR-10}
 & U & $57.91{\pm}0.40$ & $55.53{\pm}0.26$ & $50.99{\pm}0.28$ & $45.15{\pm}0.26$ & $42.95{\pm}1.15$ \\
 & G & $57.83{\pm}0.35$ & $55.56{\pm}0.14$ & $52.44{\pm}0.18$ & $50.32{\pm}0.33$ & $49.02{\pm}0.06$ \\
\midrule
\multirow{2}{*}{CIFAR-100}
 & U & $24.83{\pm}0.18$ & $21.26{\pm}0.12$ & $17.00{\pm}0.10$ & $10.53{\pm}0.52$ & $8.24{\pm}0.34$ \\
 & G & $24.93{\pm}0.27$ & $21.73{\pm}0.27$ & $18.46{\pm}0.28$ & $15.89{\pm}0.34$ & $14.72{\pm}0.20$ \\
\midrule
\multirow{2}{*}{Tiny-IN}
 & U & $10.44{\pm}0.17$ & $9.02{\pm}0.18$ & $7.27{\pm}0.32$ & $5.39{\pm}0.17$ & $4.61{\pm}0.13$ \\
 & G & $10.40{\pm}0.23$ & $9.19{\pm}0.23$ & $7.83{\pm}0.13$ & $6.53{\pm}0.16$ & $6.01{\pm}0.21$ \\
\bottomrule
\end{tabular}
\end{table}

\begin{table}[ht]
\centering
\scriptsize
\caption{DST top-1 accuracy ($\%$) under \textbf{uniform} initialization:
Magnitude/Random (U) vs.\ reachability-guided (G).}
\label{tab:app-unif}
\begin{tabular}{llccccc}
\toprule
Dataset & & $99\%$ & $99.5\%$ & $99.75\%$ & $99.875\%$ & $99.9\%$ \\
\midrule
\multirow{2}{*}{CIFAR-10}
 & U & $54.79{\pm}0.38$ & $48.50{\pm}3.53$ & $30.78{\pm}3.68$ & $16.55{\pm}5.09$ & $16.08{\pm}1.85$ \\
 & G & $57.43{\pm}0.35$ & $55.35{\pm}0.34$ & $40.14{\pm}3.72$ & $25.93{\pm}5.71$ & $22.99{\pm}0.96$ \\
\midrule
\multirow{2}{*}{CIFAR-100}
 & U & $22.38{\pm}0.46$ & $15.13{\pm}0.64$ & $7.83{\pm}0.43$ & $4.15{\pm}0.28$ & $3.17{\pm}0.73$ \\
 & G & $23.22{\pm}0.13$ & $19.98{\pm}0.43$ & $14.48{\pm}0.60$ & $9.55{\pm}0.22$ & $8.03{\pm}0.42$ \\
\midrule
\multirow{2}{*}{Tiny-IN}
 & U & $8.17{\pm}0.23$ & $6.46{\pm}0.28$ & $3.66{\pm}0.63$ & $1.94{\pm}0.17$ & $1.77{\pm}0.31$ \\
 & G & $8.39{\pm}0.13$ & $7.53{\pm}0.27$ & $6.05{\pm}0.31$ & $4.37{\pm}0.22$ & $3.67{\pm}0.21$ \\
\bottomrule
\end{tabular}
\end{table}

\paragraph{MLP-Mixer efficiency.}
Table~\ref{tab:app-mixer} repeats the Mixer comparison with top-3 accuracy,
training FLOPs ($\times 10^{12}$), and active parameters ($\times 10^{3}$).
Guided Random matches or exceeds gradient regrowth at $99.95\%$ sparsity at
essentially the same active-parameter budget as random regrowth, without requiring
gradient-based regrowth.

\begin{table}[ht]
\centering
\scriptsize
\caption{MLP-Mixer (CIFAR-100, ER): top-3 accuracy, training FLOPs ($\times 10^{12}$),
and active parameters ($\times 10^{3}$).}
\label{tab:app-mixer}
\begin{tabular}{lcccc}
\toprule
Sparsity / method & Acc@3\,(\%) & Loss & FLOPs & Params \\
\midrule
$0\%$ Dense & $66.33{\pm}1.11$ & $1.28{\pm}0.03$ & $16300$ & $21700$ \\
\midrule
$99.00\%$ Mag./Grad. & $71.92{\pm}1.70$ & $2.39{\pm}0.09$ & $18.7$ & $216.8$ \\
$99.00\%$ Mag./Rand. & $68.73{\pm}0.85$ & $2.52{\pm}0.04$ & $18.5$ & $217.1$ \\
$99.00\%$ G-Mag./Rand. & $68.12{\pm}0.45$ & $2.66{\pm}0.03$ & $18.6$ & $217.1$ \\
\midrule
$99.90\%$ Mag./Grad. & $61.08{\pm}0.81$ & $3.15{\pm}0.02$ & $4.20$ & $21.7$ \\
$99.90\%$ Mag./Rand. & $45.81{\pm}0.72$ & $3.60{\pm}0.01$ & $4.02$ & $21.7$ \\
$99.90\%$ G-Mag./Rand. & $57.48{\pm}0.86$ & $3.25{\pm}0.03$ & $4.03$ & $21.8$ \\
\midrule
$99.95\%$ Mag./Grad. & $42.31{\pm}2.89$ & $3.68{\pm}0.08$ & $3.39$ & $10.9$ \\
$99.95\%$ Mag./Rand. & $31.96{\pm}1.11$ & $3.98{\pm}0.02$ & $3.21$ & $10.9$ \\
$99.95\%$ G-Mag./Rand. & $47.65{\pm}0.76$ & $3.53{\pm}0.02$ & $3.22$ & $10.9$ \\
\midrule
$99.99\%$ Mag./Grad. & $13.28{\pm}3.59$ & $4.39{\pm}0.07$ & $2.75$ & $2.20$ \\
$99.99\%$ Mag./Rand. & $4.22{\pm}1.21$ & $4.60{\pm}0.01$ & $2.57$ & $2.16$ \\
$99.99\%$ G-Mag./Rand. & $13.56{\pm}1.70$ & $4.42{\pm}0.03$ & $2.57$ & $2.17$ \\
\bottomrule
\end{tabular}
\end{table}

%% file: refs.bib
@article{erdos1960evolution,
  title   = {On Random Graphs~{I}},
  author  = {Erd{\H{o}}s, Paul and R{\'e}nyi, Alfr{\'e}d},
  journal = {Publicationes Mathematicae Debrecen},
  volume  = {6},
  pages   = {290--297},
  year    = {1959}
}

@inproceedings{evci2020rigging,
  title     = {Rigging the Lottery: Making All Tickets Winners},
  author    = {Evci, Utku and Gale, Trevor and Menick, Jacob and Castro, Pablo Samuel and Elsen, Erich},
  booktitle = {Proceedings of the 37th International Conference on Machine Learning},
  series    = {Proceedings of Machine Learning Research},
  volume    = {119},
  pages     = {2943--2951},
  year      = {2020},
  publisher = {PMLR}
}

@article{
franssen2025connect,
title={Co{NN}ect: Connectivity-Based Regularization for Structural Pruning of Neural Networks},
author={Christian P.C. Franssen and Jinyang Jiang and Yijie Peng and Bernd Heidergott},
journal={Transactions on Machine Learning Research},
issn={2835-8856},
year={2025},
url={https://openreview.net/forum?id=RIZCe7BuEp},
note={}
}

@article{hoefler2021sparsity,
  title   = {Sparsity in Deep Learning: Pruning and Growth for Efficient Inference and Training in Neural Networks},
  author  = {Hoefler, Torsten and Alistarh, Dan and Ben-Nun, Tal and Dryden, Nikoli and Peste, Alexandra},
  journal = {Journal of Machine Learning Research},
  volume  = {22},
  number  = {241},
  pages   = {1--124},
  year    = {2021}
}

@inproceedings{ji2024advancing,
  title     = {Advancing Dynamic Sparse Training by Exploring Optimization Opportunities},
  author    = {Ji, Jie and Li, Gen and Yin, Lu and Qin, Minghai and Yuan, Geng and Guo, Linke and Liu, Shiwei and Ma, Xiaolong},
  booktitle = {Proceedings of the 41st International Conference on Machine Learning},
  series    = {Proceedings of Machine Learning Research},
  volume    = {235},
  pages     = {21606--21619},
  year      = {2024},
  publisher = {PMLR}
}

@article{li2025pushing,
  title   = {Pushing the Limits of Sparsity: A Bag of Tricks for Extreme Pruning},
  author  = {Li, Andy and Durrant, Aiden and Markovic, Milan and Huang, Tianjin and Kundu, Souvik and Chen, Tianlong and Yin, Lu and Leontidis, Georgios},
  journal = {Transactions on Machine Learning Research},
  year    = {2025},
  url     = {https://openreview.net/forum?id=XX9JdOJD8R}
}

@inproceedings{liu2020topological,
  title     = {Topological Insights into Sparse Neural Networks},
  author    = {Liu, Shiwei and {Van der Lee}, Tim and Yaman, Anil and Atashgahi, Zahra and Ferraro, Davide and Sokar, Ghada and Pechenizkiy, Mykola and Mocanu, Decebal Constantin},
  booktitle = {Machine Learning and Knowledge Discovery in Databases},
  volume    = {12977},
  pages     = {279--294},
  year      = {2021},
  publisher = {Springer International Publishing},
}

@inproceedings{liu2021itop,
  title     = {Do We Actually Need Dense Over-Parameterization? In-Time Over-Parameterization in Sparse Training},
  author    = {Liu, Shiwei and Yin, Lu and Mocanu, Decebal Constantin and Pechenizkiy, Mykola},
  booktitle = {Proceedings of the 38th International Conference on Machine Learning},
  series    = {Proceedings of Machine Learning Research},
  volume    = {139},
  pages     = {6989--7000},
  year      = {2021},
  publisher = {PMLR}
}

@article{mocanu2018scalable,
  title   = {Scalable Training of Artificial Neural Networks with Adaptive Sparse Connectivity Inspired by Network Science},
  author  = {Mocanu, Decebal Constantin and Mocanu, Elena and Stone, Peter and Nguyen, Phuong H. and Gibescu, Madeleine and Liotta, Antonio},
  journal = {Nature Communications},
  volume  = {9},
  number  = {1},
  pages   = {2383},
  year    = {2018}
}

@inproceedings{nowak2023fantastic,
  title     = {Fantastic Weights and How to Find Them: Where to Prune in Dynamic Sparse Training},
  author    = {Nowak, Aleksandra I. and Grooten, Bram and Mocanu, Decebal Constantin and Tabor, Jacek},
  booktitle = {Advances in Neural Information Processing Systems},
  volume    = {36},
  pages     = {55160--55192},
  year      = {2023}
}

@inproceedings{tanaka2020pruning,
  title     = {Pruning Neural Networks without Any Data by Iteratively Conserving Synaptic Flow},
  author    = {Tanaka, Hidenori and Kunin, Daniel and Yamins, Daniel L. K. and Ganguli, Surya},
  booktitle = {Advances in Neural Information Processing Systems},
  volume    = {33},
  year      = {2020}
}

@techreport{krizhevsky2009learning,
  author      = {Krizhevsky, Alex},
  title       = {Learning Multiple Layers of Features from Tiny Images},
  institution = {University of Toronto},
  year        = {2009}
}

@inproceedings{le2015tiny,
  author    = {Le, Ya and Yang, Xuan},
  title     = {Tiny {ImageNet} Visual Recognition Challenge},
  booktitle = {CS231N Course Report, Stanford University},
  year      = {2015}
}

@inproceedings{tolstikhin2021mlp,
  author    = {Tolstikhin, Ilya O. and Houlsby, Neil and Kolesnikov, Alexander and Beyer, Lucas and Zhai, Xiaohua and Unterthiner, Thomas and Yung, Jessica and Steiner, Andreas and Keysers, Daniel and Uszkoreit, Jakob and Lucic, Mario and Dosovitskiy, Alexey},
  title     = {{MLP-Mixer}: An All-{MLP} Architecture for Vision},
  booktitle = {Advances in Neural Information Processing Systems},
  volume    = {34},
  pages     = {24261--24272},
  year      = {2021}
}

@mastersthesis{Jaworski2026,
month = {June},
author = {Jaworski, M.Z.},
year= {2026},
school= {University of Twente},
address= {Enschede},
type= {Thesis},
title= {When Sparsity Breaks Connectivity: Restoring Functional Paths in Neural Networks},
url= { https://purl.utwente.nl/essays/110108},
}
